\documentclass[11pt]{article}

\usepackage{epsfig}
\usepackage{theorem}
\usepackage{enumerate}
\usepackage{amsmath}
\usepackage{amssymb}
\usepackage{boxedminipage}
\usepackage{subfigure}
\usepackage{url}
\usepackage{color}
\usepackage{epsfig}

\newcommand{\titlestring}{Analysis of Estimation of Distribution Algorithms and Genetic Algorithms on NK Landscapes}
\newcommand{\reportnumber}{2008001}
\newcommand{\shortauthors}{Martin Pelikan}
\newcommand{\datestring}{January 2008}
\definecolor{myblue}{rgb}{0.165,0.34,0.5}
\date{}

\begin{document}

\begin{titlepage}
\setlength{\parindent}{0pt}
%\vspace*{0.5in}

\noindent
\includegraphics[width=5in]{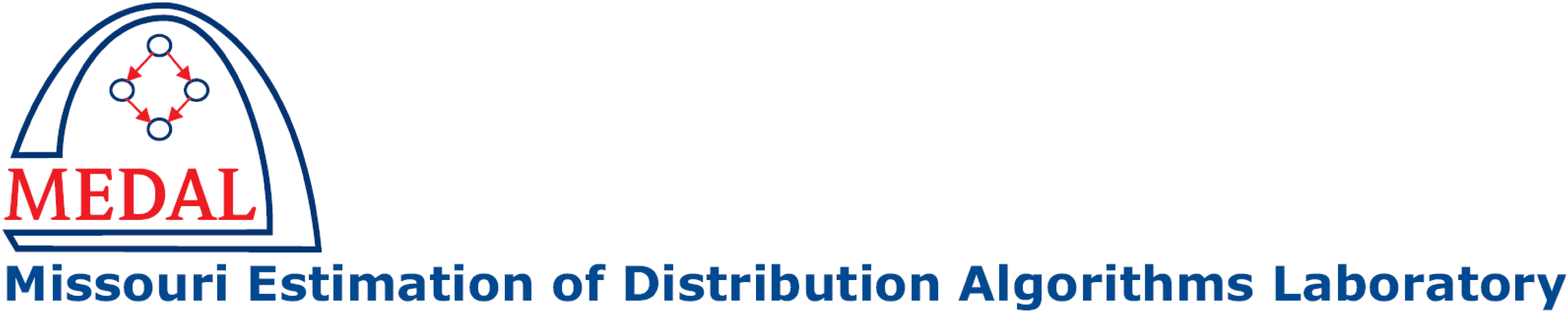}
\vspace*{0.075in}
{\color{myblue}
\hrule height 2pt
}
\vspace*{0.5in}

{\bf
\textsf{{\large
\titlestring}}
}
\vspace*{0.25in}

\textsf{\shortauthors}

\vspace*{0.25in}

\textsf{MEDAL Report No. \reportnumber}

\vspace*{0.25in}

\textsf{\datestring}

\vspace*{0.25in}

{\bf \textsf{Abstract}}  

\vspace*{0.075in}

{\small \textsf{This study analyzes performance of several genetic and evolutionary algorithms on randomly generated NK fitness landscapes with various values of $n$ and $k$. A large number of NK problem instances are first generated for each $n$ and $k$, and the global optimum of each instance is obtained using the branch-and-bound algorithm. Next, the hierarchical Bayesian optimization algorithm (hBOA), the univariate marginal distribution algorithm (UMDA), and the simple genetic algorithm (GA) with uniform and two-point crossover operators are applied to all generated instances. Performance of all  algorithms is then analyzed and compared, and the results are discussed.}}
%\vspace{4in}

\vspace*{0.25in}

{\bf \textsf{Keywords}}

\vspace*{0.075in}
{\small \textsf{NK fitness landscape, hierarchical BOA, genetic algorithm, branch and bound, performance analysis, scalability, local search, crossover.}}

\vfill

\noindent
\begin{minipage}{6in}
%\centering
{\small \textsf{Missouri Estimation of Distribution Algorithms Laboratory (MEDAL)\\
Department of Mathematics and Computer Science\\
University of Missouri--St. Louis\\
One University Blvd.,
St. Louis, MO 63121\\
E-mail: \url{medal@cs.umsl.edu}\\
WWW: \url{http://medal.cs.umsl.edu/}\\}}
\end{minipage}

\end{titlepage}

\title{\titlestring}

\author{
{\bf Martin Pelikan}\\
Missouri Estimation of Distribution Algorithms Laboratory (MEDAL)\\
Dept. of Math and Computer Science, 320 CCB\\
University of Missouri at St. Louis\\
One University Blvd., St. Louis, MO 63121\\
\url{pelikan@cs.umsl.edu}}

\maketitle

%==============================================================

\begin{abstract}
This study analyzes performance of several genetic and evolutionary algorithms on randomly generated NK fitness landscapes with various values of $n$ and $k$. A large number of NK problem instances are first generated for each $n$ and $k$, and the global optimum of each instance is obtained using the branch-and-bound algorithm. Next, the hierarchical Bayesian optimization algorithm (hBOA), the univariate marginal distribution algorithm (UMDA), and the simple genetic algorithm (GA) with uniform and two-point crossover operators are applied to all generated instances. Performance of all  algorithms is then analyzed and compared, and the results are discussed.
\end{abstract}

\noindent
{\bf Keywords:} NK fitness landscape, hierarchical BOA, genetic algorithm, branch and bound, performance analysis, scalability, local search, crossover.

%==============================================================

\section{Introduction}
NK fitness landscapes~\cite{Kauffman:89,Kauffman:93} were introduced by Kauffman as tunable models of rugged fitness landscape. An NK landscape is a function defined on binary strings of fixed length and is characterized by two parameters: (1)~$n$ for the overall number of bits and (2)~$k$ for the neighborhood size. For each bit, $k$ neighbors are specified and a function is given that determines the fitness contribution of the bit and its neighbors. Usually, the function for each bit is given as a lookup table of size $2^{k+1}$ (one value for each combination of the bit and its neighbors), and both the neighbors as well as the subfunction lookup tables are initialized randomly in some way. 

NK landscapes are NP-complete for $k>1$, although some variants of NK landscapes are polynomially solvable and there exist approximation algorithms for other cases~\cite{Wright:00,Gao:02,Choi:05}. Nonetheless, NK landscapes remain a challenge for any optimization algorithm and they are also interesting from the perspective of complexity theory and computational biology; that is why since their inception NK landscapes have attracted researchers in all these areas~\cite{Kauffman:93,altenberg:97,Wright:00,Gao:02,Aguirre:03,Choi:05}. 

This paper presents an in-depth empirical performance analysis of various genetic and evolutionary algorithms on NK landscapes with varying $n$ and $k$. For each value of $n$ and $k$, a large number of problem instances are first generated. Then, the branch-and-bound algorithm is applied to each of these instances to provide a guaranteed global optimum of this instance. Although the application of branch and bound limits the size of problems that we can study, one of the primary goals was to ensure that we are able to verify the global optimum of each instance for every algorithm considered in this study. Several genetic and evolutionary algorithms are then applied to all generated problem instances and their performance is analyzed and compared. More specifically, we consider the hierarchical Bayesian optimization algorithm (hBOA), the univariate marginal distribution algorithm (UMDA), and the simple genetic algorithm (GA) with bit-flip mutation, and uniform or two-point crossover operator. Additionally, GA without any crossover is considered. The results provide insight into the difficulty of NK landscapes with respect to the parameters $n$ and $k$ and performance differences between all compared algorithms. Several interesting avenues for future work are outlined. 

The paper starts by describing NK landscapes and the branch-and-bound algorithm used to verify the global optima of generated NK landscapes in section~\ref{section-landscapes}. Section~\ref{section-algorithms} outlines compared algorithms. Section~\ref{section-experiments} presents experimental results. Section~\ref{section-future-work} discusses future work. Finally, section~\ref{section-conclusions} summarizes and concludes the paper. 

%==============================================================

\section{NK Landscapes}
\label{section-landscapes}
This section describes NK landscapes and a method to generate random problem instances of NK landscapes. Additionally, the section describes the branch-and-bound algorithm, which was used to obtain global optima of all NK problem instances considered in this paper. Branch and bound is a complete algorithm and it is thus guaranteed to find the true global optimum; this was especially useful for scalability experiments and performance analyses of different evolutionary algorithms. Nonetheless, branch and bound requires exponential time and thus the size of instances it can solve in practical time is severely limited. 

\subsection{Problem Definition}
An NK fitness landscape~\cite{Kauffman:89,Kauffman:93} is fully defined by the following components:
\begin{itemize}
\item The number of bits, $n$.
\item The number of neighbors per bit, $k$. 
\item A set of $k$ neighbors $\Pi(X_i)$ for the $i$-th bit, $X_i$, for every $i\in\{0,\ldots, n-1\}$. 
\item A subfunction $f_i$ defining a real value for each combination of values of $X_i$ and $\Pi(X_i)$ for every $i\in\{0,\ldots, n-1\}$. Typically, each subfunction is defined as a lookup table with $2^{k+1}$ values. 
\end{itemize}
The objective function $f_{nk}$ to maximize is then defined as 
\[
f_{nk}(X_0,X_1,\ldots, X_{n-1}) = \sum_{i=0}^{n-1} f_i(X_i,\Pi(X_i)).
\]

The difficulty of optimizing NK landscapes depends on all of the four components defining an NK problem instance. One useful approach to analyzing complexity of NK landscapes is to focus on the influence of $k$ on problem complexity. For $k=0$, NK landscapes are simple unimodal functions similar to onemax or binint, which can be solved in linear time and should be easy for practically any genetic and evolutionary algorithm.  The global optimum of NK landscapes can be obtained in polynomial time~\cite{Wright:00} even for $k=1$; on the other hand, for $k>1$, the problem of finding the global optimum of unrestricted NK landscapes is NP-complete~\cite{Wright:00}. The problem becomes polynomially solvable with dynamic programming even for $k>1$ if the neighbors are restricted to only adjacent string positions (using circular strings)~\cite{Wright:00} or if the subfunctions are generated according to some distributions~\cite{Gao:02}. For unrestricted NK landscapes with $k>1$, a polynomial-time approximation algorithm exists with the approximation threshold $1-1/2^{k+1}$~\cite{Wright:00}.

\subsection{Generating Random NK Problem Instances}
Typically, both the neighbors as well as the lookup tables defining the subfunctions are generated randomly. In this paper, for each string position $X_i$, we first generate a random set of $k$ neighbors where each string position except for $X_i$ is selected with equal probability. Then, the lookup table defining $f_i$ is generated using the uniform distribution over $[0,1)$. 

Consequently, the studied class of NK landscapes is NP-complete for any $k>1$. Since the case for $k=1$ is extremely simple to solve, we only considered $k>1$; specifically, we considered $k=2$ to $6$ with step $1$. To study scalability of various evolutionary algorithms, for each $k$, we considered a range of values of $n$ with the minimum value of $n=20$ and the maximum value bounded mainly by the available computational resources and the scope of the empirical analysis.

\subsection{Branch and Bound}
\begin{figure}
\centering
\epsfig{file=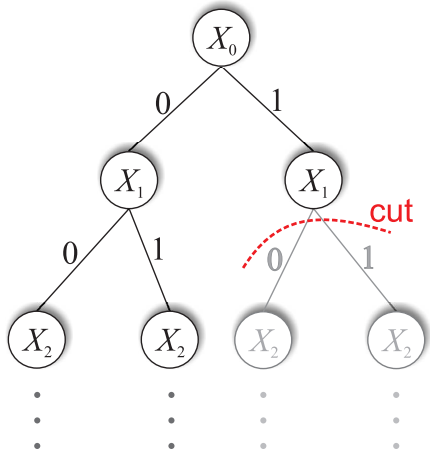,width=2.25in}
\caption{Branch and bound traverses the recursion tree where each each level sets the value of one bit and each leaf thus corresponds to one instance of all $n$ bits. Subtrees that lead to solutions that cannot improve the best-so-far solution are cut to improve efficiency.}
\label{fig-bb-procedure}
\end{figure}

The basic idea of branch and bound is to recursively explore all possible binary strings of $n$ bits using a recursion tree where each level corresponds to one of the bits and the subtrees below each level correspond to the different values of the bit corresponding to this level. To make the algorithm more efficient, some subtrees are cut if they can be proven to not lead to any solution that is better than the best-so-far solution found. While this cannot eliminate the exponential complexity, which can be expected due to the NP-completeness for NK landscapes with $k>1$, it significantly improves the performance of the algorithm and allows it to solve much larger problem instances than if a complete recursion tree had to be explored. The branch-and-bound procedure is illustrated in figure~\ref{fig-bb-procedure}.

Before running the branch-and-bound algorithm, we first use a simple hill climber based on bit-flip mutation with several random restarts to locate high-quality local optima. The best of the discovered optima is then used as the best-so-far solution when the branch-and-bound algorithm is started. In the branch-and-bound approach used in this paper, the bits are assigned sequentially from $X_0$ to $X_n$ (there are two subtrees of each node at level $i$, each corresponding to one value of $X_i$), although reordering the bits might improve performance under some conditions. 

When processing a node at level $i$, the best value we can obtain by setting the remaining $n-i$ bits is given by
\[
\max_{x_i,\ldots,x_n\in\{0,1\}^{n-i}} f_{nk}(X_1=x_1,\ldots,X_n=x_n)
\]
where bits $x_0$ to $x_{i-1}$ are assumed to be fixed to the values defined by the path from the root of the recursion tree to the current node. If a solution has been found already that has a higher fitness than this maximum possible value, the processing below the currently processed node does not have to continue and the remaining unexplored parts of the recursion tree can be explored with the exception of those parts that have already been cut. 

We also tried another variant of the branch-and-bound algorithm, in which the best value of the objective function is computed incrementally for subsets containing only the first $i$ bits with $i=2$ to $n$. While this approach has been very efficient in solving instances of the Sherrington-Kirkpatrick spin glass model~\cite{Kobe:84}, for NK landscapes, the algorithm described earlier performed more efficiently. 

The aforedescribed branch-and-bound algorithm is complete and it is thus guaranteed to find the global optimum of any problem instance. Nonetheless, the complexity of branch and bound can be expected to grow exponentially fast and solving large NK instances becomes intractable with this algorithm. For example, for $k=2$, the proposed branch-and-bound algorithm was fast enough to solve ten thousand unique instances of $n\leq 52$; for $k=6$, the algorithm was fast enough to deal with instances of size $n\leq 36$. While the evolutionary algorithms presented in the next section should be capable of reliably solving larger instances, their convergence to the global optimum cannot be guaranteed; nonetheless, section~\ref{section-future-work} discusses how to extend this study to deal with larger NK problem instances, which are intractable with the branch-and-bound algorithm.

%==============================================================

\section{Compared Algorithms}
\label{section-algorithms}
This section outlines the optimization algorithms discussed in this paper: (1) the hierarchical Bayesian optimization algorithm (hBOA)~\cite{Pelikan:01*,Pelikan:book}, (2)~the univariate marginal distribution algorithm (UMDA)~\cite{Muhlenbein:96**}, and (3)~the genetic algorithm (GA)~\cite{Holland:75a,Goldberg:89d}. Additionally, the section describes the deterministic hill climber (DHC)~\cite{Pelikan:03*}, which is incorporated into all compared algorithms to improve their performance. In all compared algorithms, candidate solutions are represented by binary strings of $n$ bits and a niching technique called restricted tournament replacement (RTR)~\cite{Harik:95a} is used for effective diversity maintenance. 

\subsection{Genetic Algorithm}
The genetic algorithm (GA)~\cite{Holland:75a,Goldberg:89d} evolves a population of candidate solutions typically represented by fixed-length binary strings. The first population is generated at random. Each iteration starts by selecting promising solutions from the current population. We use binary tournament selection. New solutions are created by applying variation operators to the population of selected solutions. Specifically, crossover is used to exchange bits and pieces between pairs of candidate solutions and mutation is used to perturb the resulting solutions. Here we use uniform or two-point crossover, and bit-flip mutation~\cite{Goldberg:89d}. To ensure effective diversity maintenance, the new candidate solutions are incorporated into the original population using restricted tournament replacement (RTR)~\cite{Harik:95a}. The run is terminated when termination criteria are met.

\subsection{Univariate Marginal Distribution Algorithm (UMDA)}
The univariate marginal distribution algorithm (UMDA)~\cite{Muhlenbein:96**} also evolves a population of candidate solutions represented by fixed-length binary strings with the initial population generated at random. Each iteration starts by selecting a population of promising solutions using any common selection method of genetic and evolutionary algorithms; we use binary tournament selection. Then, the probability vector is learned that stores the proportion of $1$s in each position of the selected population. Each bit of a new candidate solution is then set to $1$ with the probability equal to the proportion of $1$s in this position; otherwise, the bit is set to $0$. Consequently, the variation operator of UMDA preserves the proportions of $1$s in each position while decorrelating different string positions. The new candidate solutions are incorporated into the original population using RTR. The run is terminated when termination criteria are met.

UMDA is an estimation of distribution algorithm (EDA)~\cite{Baluja:94,Muhlenbein:96**,Larranaga:02,Pelikan:02}. EDAs---also called probabilistic model-building genetic algorithms (PMBGAs)~\cite{Pelikan:02} and iterated density estimation algorithms (IDEAs)~\cite{Bosman:00*}---replace standard variation operators of genetic algorithms such as crossover and mutation by building a probabilistic model of promising solutions and sampling the built model to generate new candidate solutions. The only difference between the GA and UMDA is in the way the selected solutions are processed to generate new solutions.

\subsection{Hierarchical BOA (hBOA)}
The hierarchical Bayesian optimization algorithm (hBOA)~\cite{Pelikan:01*,Pelikan:book} is also an EDA and the basic procedure of hBOA is similar to that of the UMDA variant described earlier. However, to model promising solutions and sample new solutions, Bayesian networks with local structures~\cite{Chickering:97,Friedman:99} are used instead of the simple probability vector of UMDA. Similarly as in the considered GA and UMDA variants, the new candidate solutions are incorporated into the original population using RTR and the run is terminated when termination criteria are met.

\subsection{Deterministic Hill Climber (DHC)}
The deterministic hill climber (DHC) is incorporated into GA, UMDA and hBOA to improve their performance. DHC takes a candidate solution represented by an $n$-bit binary string on input. Then, it performs one-bit changes on the solution that lead to the maximum improvement of solution quality. DHC is terminated when no single-bit flip improves solution quality and the solution is thus locally optimal. Here, DHC is used to improve every solution in the population before the evaluation is performed. 

%==============================================================

\section{Experiments}
\label{section-experiments}
This section describes experiments and presents experimental results. First, problem instances and experimental setup are discussed. Next, the analysis of hBOA, UMDA and several GA variants is presented. Finally, all algorithms are compared and the results of the comparisons are discussed.

\subsection{Problem Instances}
NK instances for $k=2$ to $k=6$ with step $1$ were studied. The only restriction on problem size was the efficiency of the branch-and-bound algorithm, the complexity of which grew very fast with $n$. For $k=2$, we considered $n=20$ to $n=52$ with step $2$; for $k=3$, we considered $n=20$ to $n=48$ with step $2$; for $k=4$, we considered $n=20$ to $n=40$ with step $2$; for $k=5$, we considered $n=20$ to $n=38$ with step $2$; finally, for $k=6$, we considered $n=20$ to $n=32$ with step $2$. 

For each combination of $n$ and $k$, we generated 10,000 random problem instances and for each instance we used the branch-and-bound algorithm to locate the global optimum. Then, we applied hBOA, UMDA and several GA variants to each of these instances and collected empirical results, which were subsequently analyzed. That means that overall 600,000 unique problem instances were generated and all of them were tested with every algorithm included in this study.

\subsection{Compared Algorithms}
The following list summarizes the algorithms included in this study:
\begin{enumerate}[(i)]
\item Hierarchical BOA (hBOA).
\item Univariate marginal distribution algorithm (UMDA).
\item Genetic algorithm with uniform crossover and bit-flip mutation.
\item Genetic algorithm with two-point crossover and bit-flip mutation.
\item Genetic algorithm with bit-flip mutation and no crossover.
\item Hill climbing (results omitted due to inferior performance and infeasible computation).
\end{enumerate}

\subsection{Experimental Setup}
To select promising solutions, binary tournament selection is used. New solutions (offspring) are incorporated into the old population using RTR with window size $w=\min\{n,N/5\}$ as suggested in ref.~\cite{Pelikan:book}. In hBOA, Bayesian networks with decision trees~\cite{Chickering:97,Friedman:99,Pelikan:book} are used and the models are evaluated using the Bayesian-Dirichlet metric with likelihood equivalence~\cite{Heckerman+al:94,Chickering:97} and a penalty for model complexity~\cite{Friedman:99,Pelikan:book}. All GA variants use bit-flip mutation with the probability of flipping each bit $p_m=1/n$. Two common crossover operators are considered in a GA: two-point and uniform crossover. For both crossover operators, the probability of applying crossover is set to $0.6$. To emphasize the importance of using crossover, the results for GA without any crossover are also included, where only bit-flip mutation is used. A stochastic hill climber with bit-flip mutation has also been considered in the initial stage, but the performance of this algorithm was far inferior compared to any other algorithm included in the comparison and it was intractable to solve most problem instances included in the comparison; that is why the results for this algorithm are omitted.

For each problem instance and each algorithm, an adequate population size is approximated with the bisection method~\cite{Sastry:01c,Pelikan:book}, which estimates the minimum population size required for reliable convergence to the optimum. Here, the bisection method finds an adequate population size for the algorithms to find the optimum in 10 out of 10 independent runs. Each run is terminated when the global optimum has been found. The results for each problem instance comprise of the following statistics: (1) the population size, (2) the number of iterations (generations), (3) the number of evaluations, and (4) the number of flips of DHC. For each value of $n$ and $k$, all observed statistics were averaged over the 10,000 random instances. Since for each instance, 10 successful runs were performed, for each $n$ and $k$  and each algorithm the results are averaged over 100,000 successful runs. Overall, for each algorithm, the results correspond to 6,000,000 successful runs on a total of 600,000 unique problem instances.

\subsection{Performance Analysis}
\begin{figure}
\begin{center}
{\epsfig{file=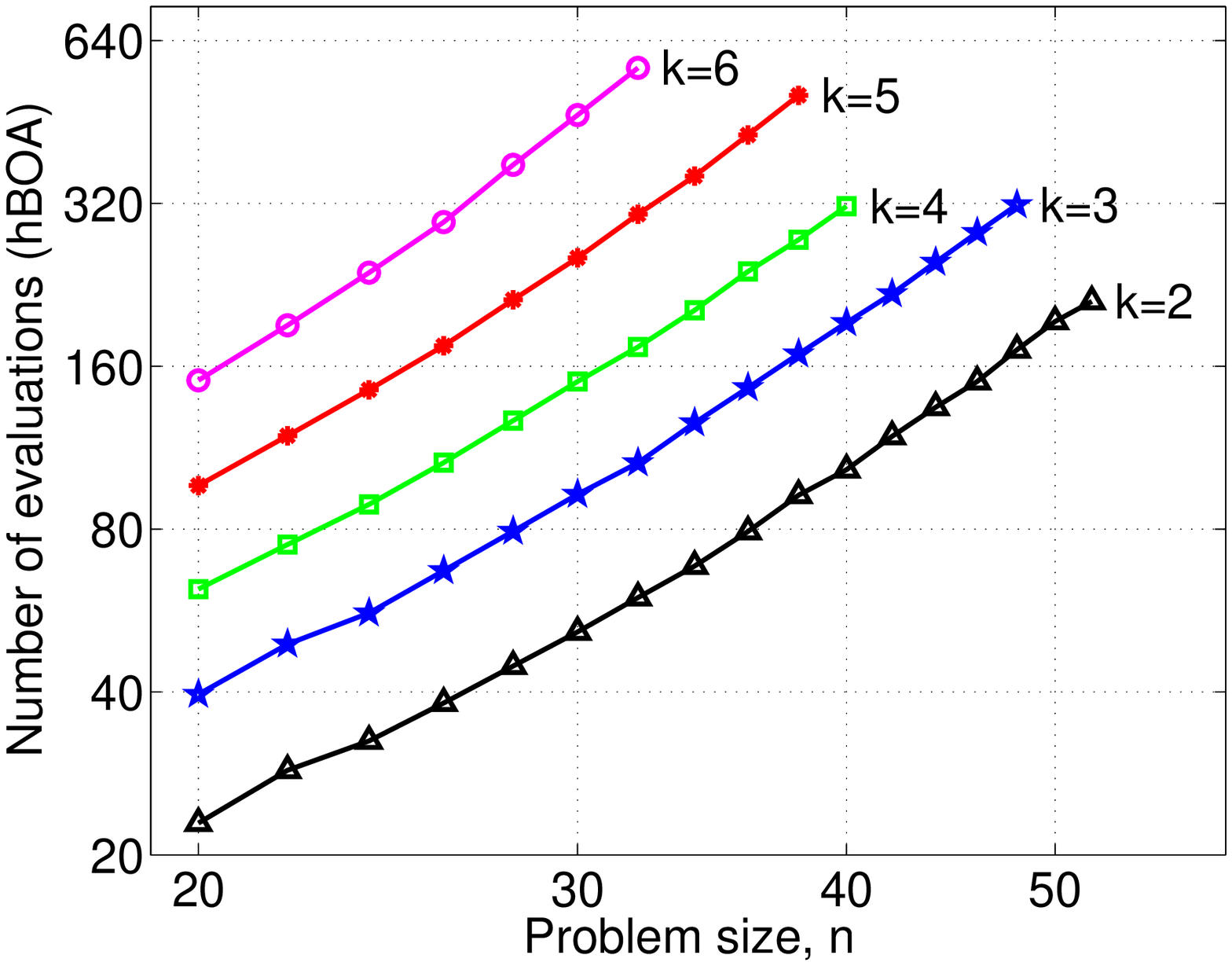,width=0.450\textwidth}}
\hspace*{3ex}
{\epsfig{file=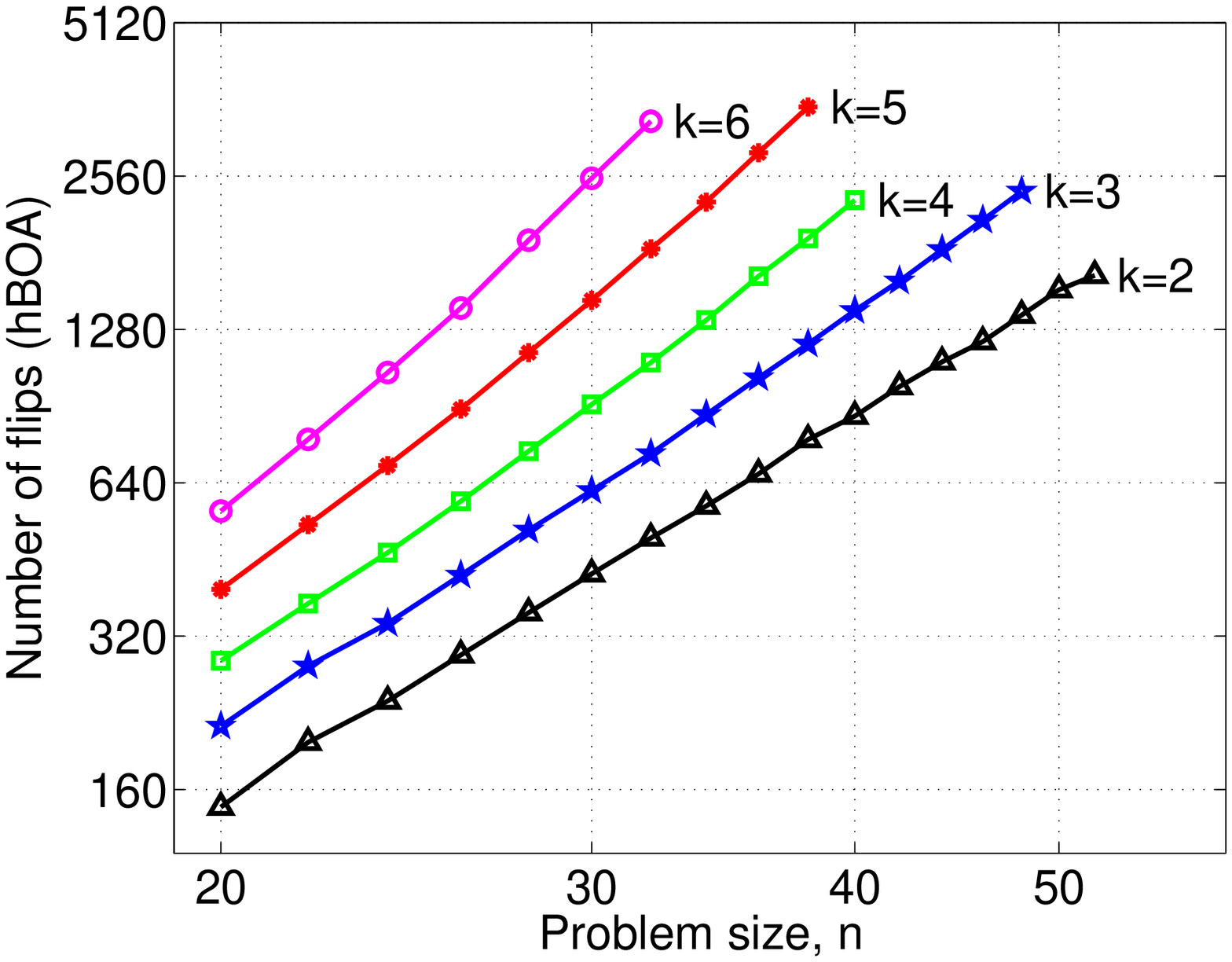,width=0.450\textwidth}}\\
{\epsfig{file=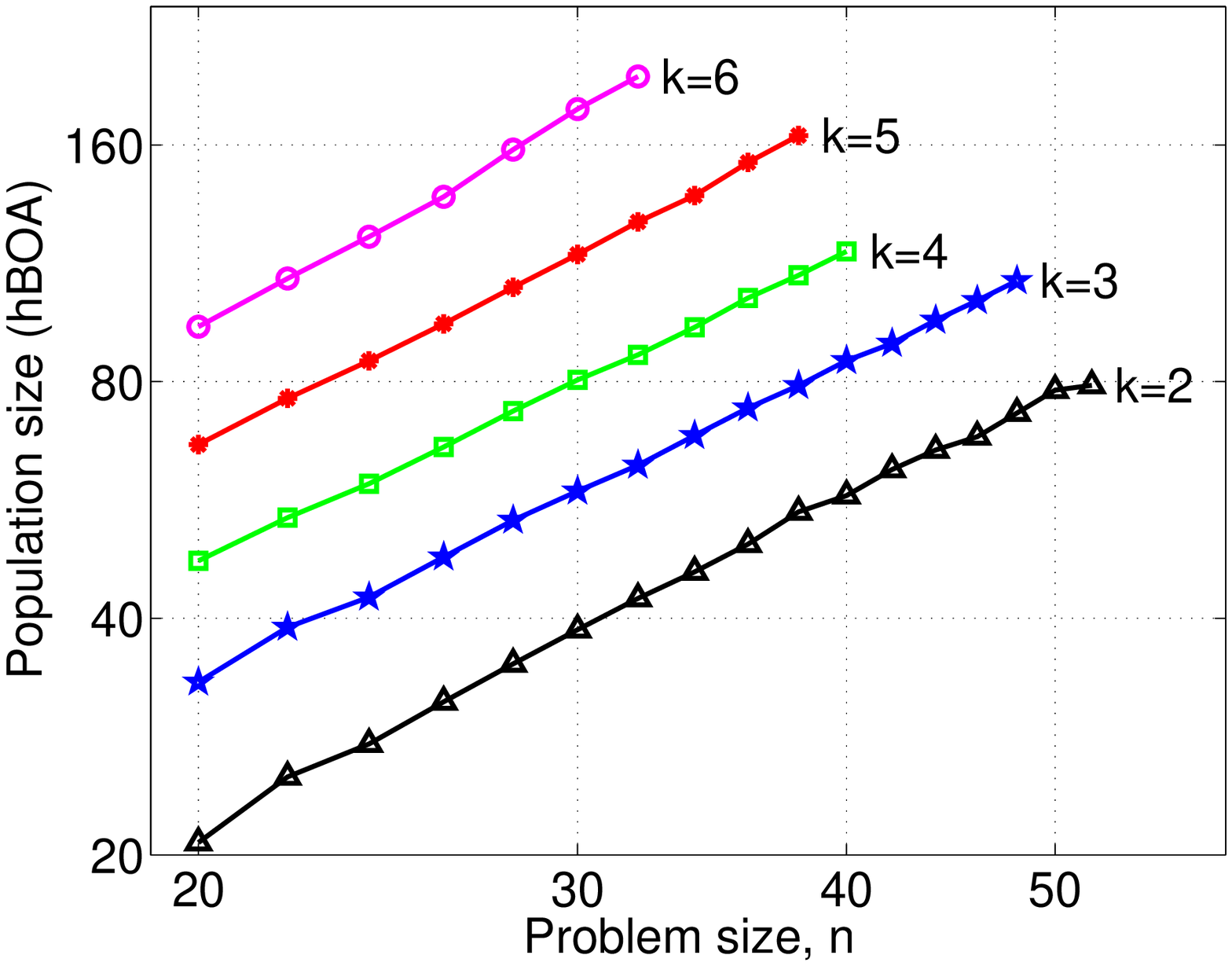,width=0.450\textwidth}}
\hspace*{3ex}
{\epsfig{file=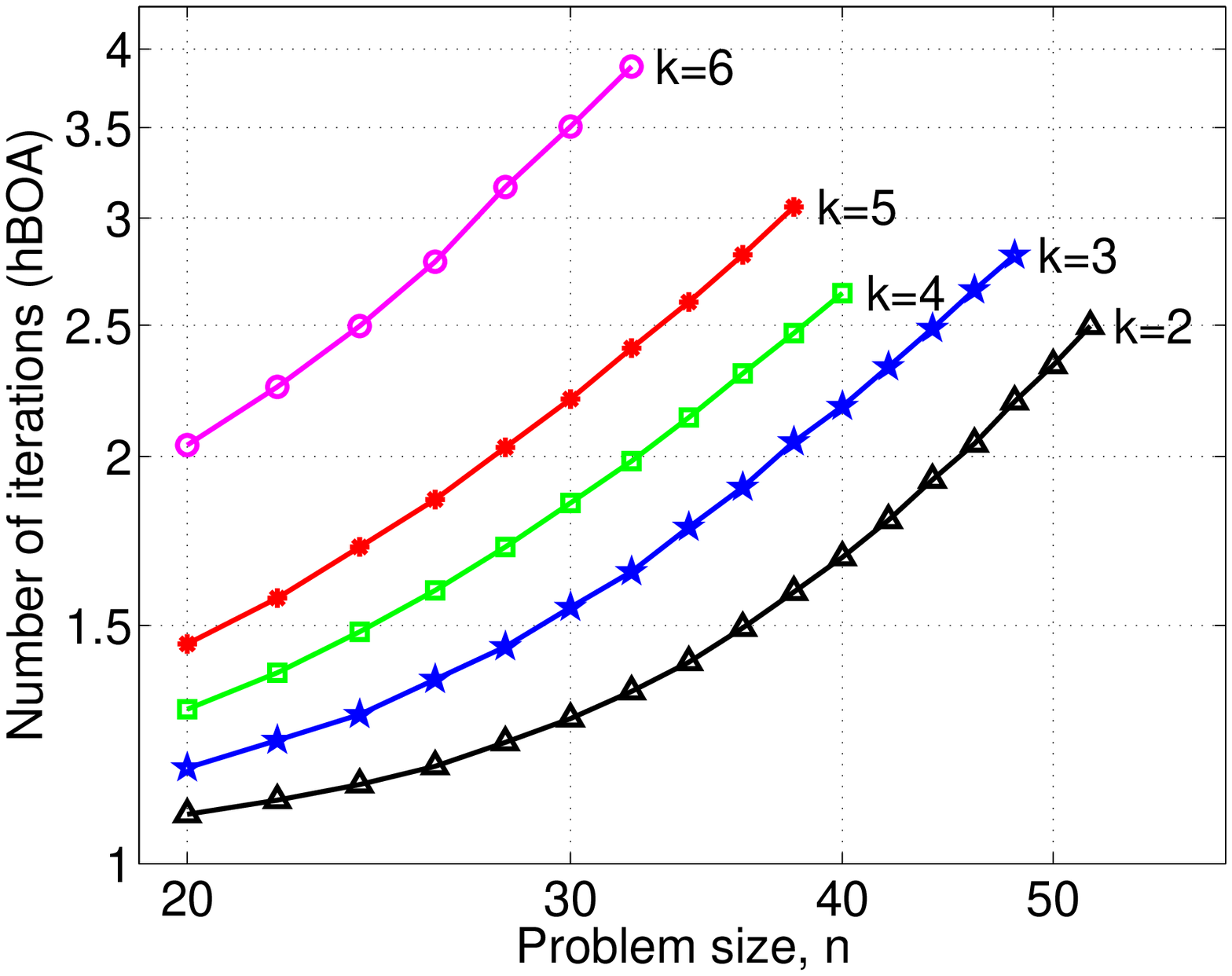,width=0.450\textwidth}}
\end{center}
\caption{Performance of hBOA on NK landscapes with $k=2$ to $6$ (log-log scale).}
\label{fig-hboa-results}
\end{figure}

Figure~\ref{fig-hboa-results} shows the average performance statistics for hBOA on NK problem instances for $k=2$ to $k=6$. As expected, performance of hBOA gets worse with increasing $k$. More specifically, the population size, the number of iterations, the number of evaluations, and the number of DHC flips appear all to grow exponentially with $k$. For a fixed $k$, the time complexity appears to grow with $n$ slightly faster than polynomially regardless of whether it is measured by the number of evaluations or the number of flips. 

\begin{figure}
\begin{center}
{\epsfig{file=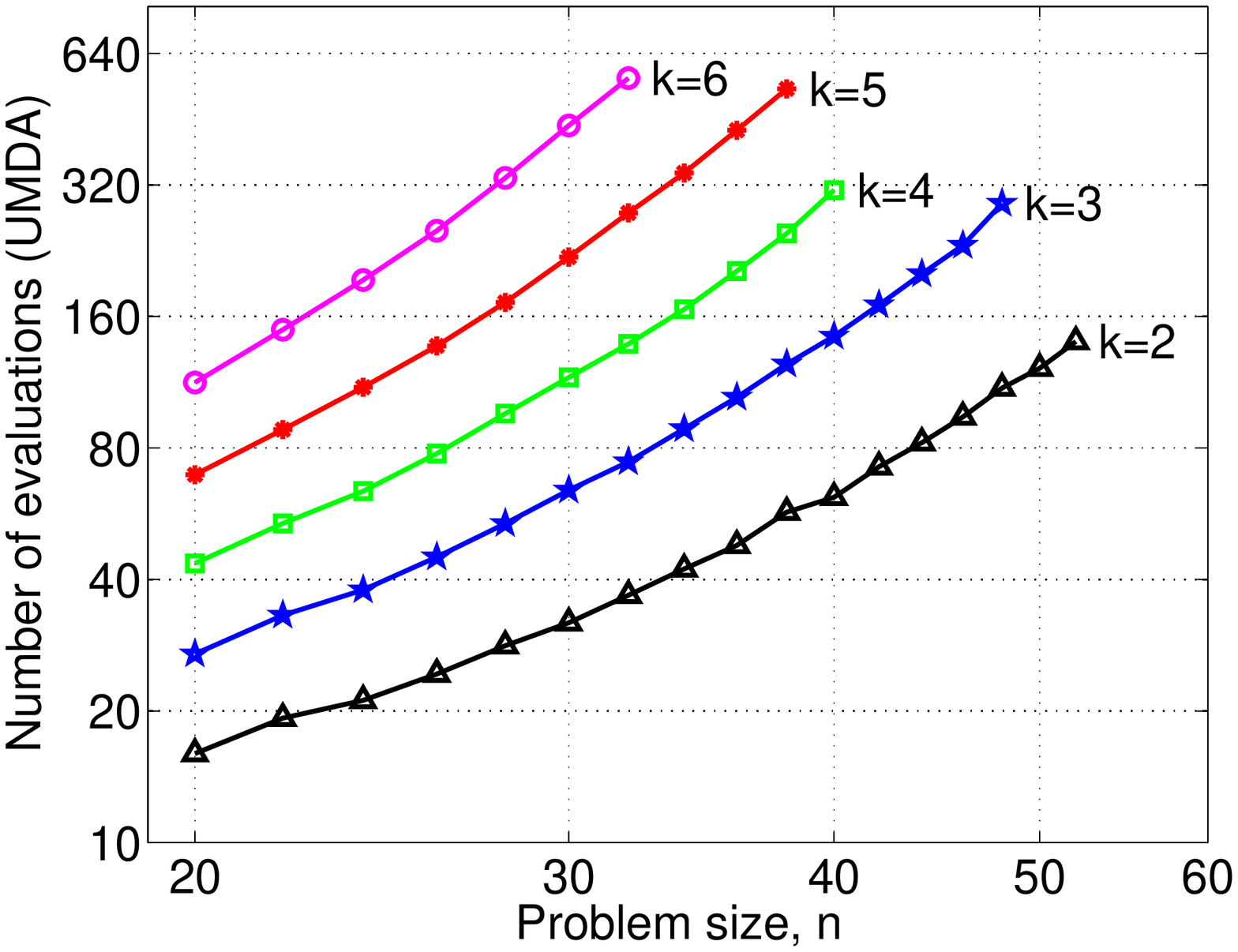,width=0.450\textwidth}}
\hspace*{3ex}
{\epsfig{file=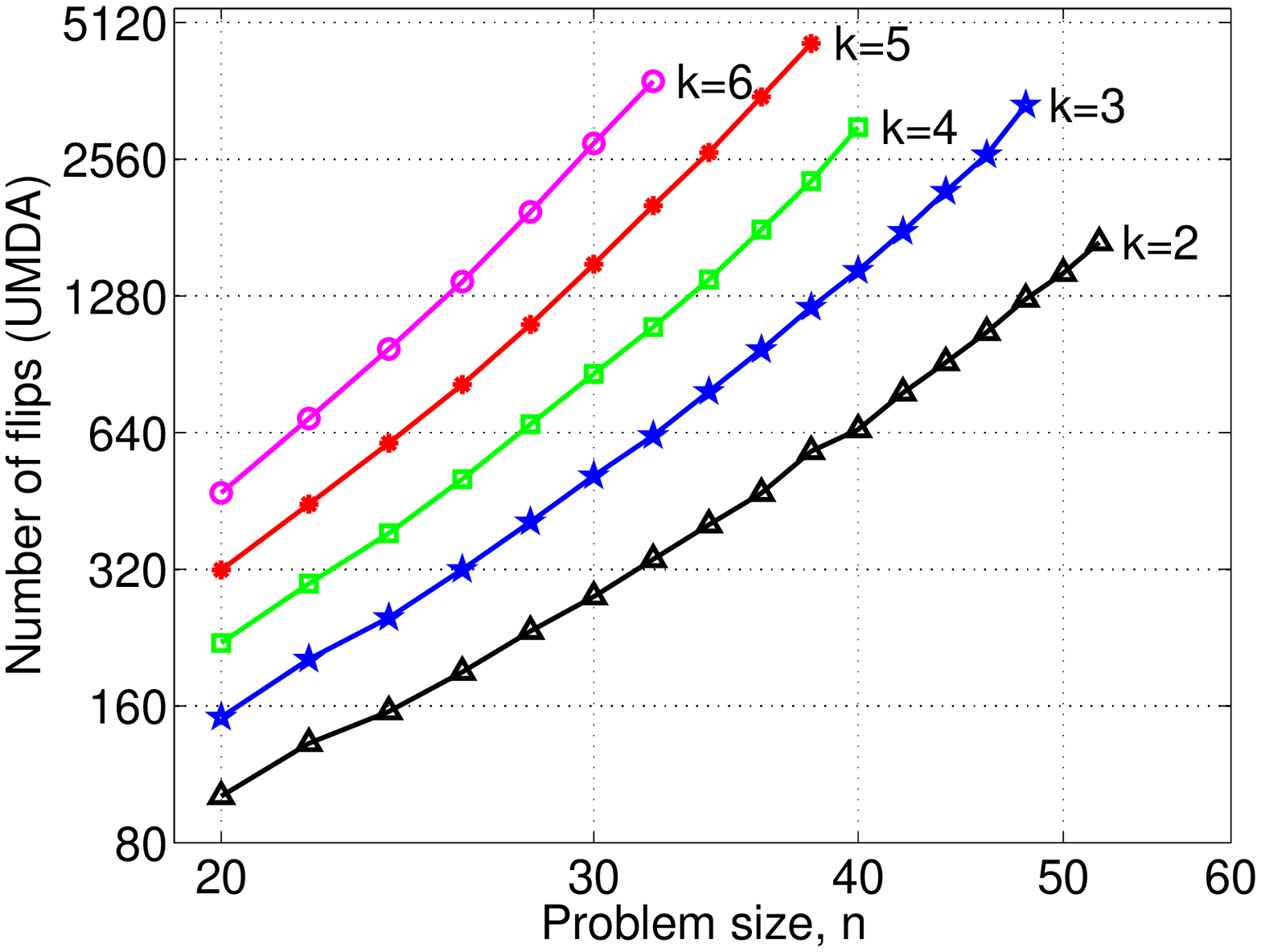,width=0.450\textwidth}}\\
{\epsfig{file=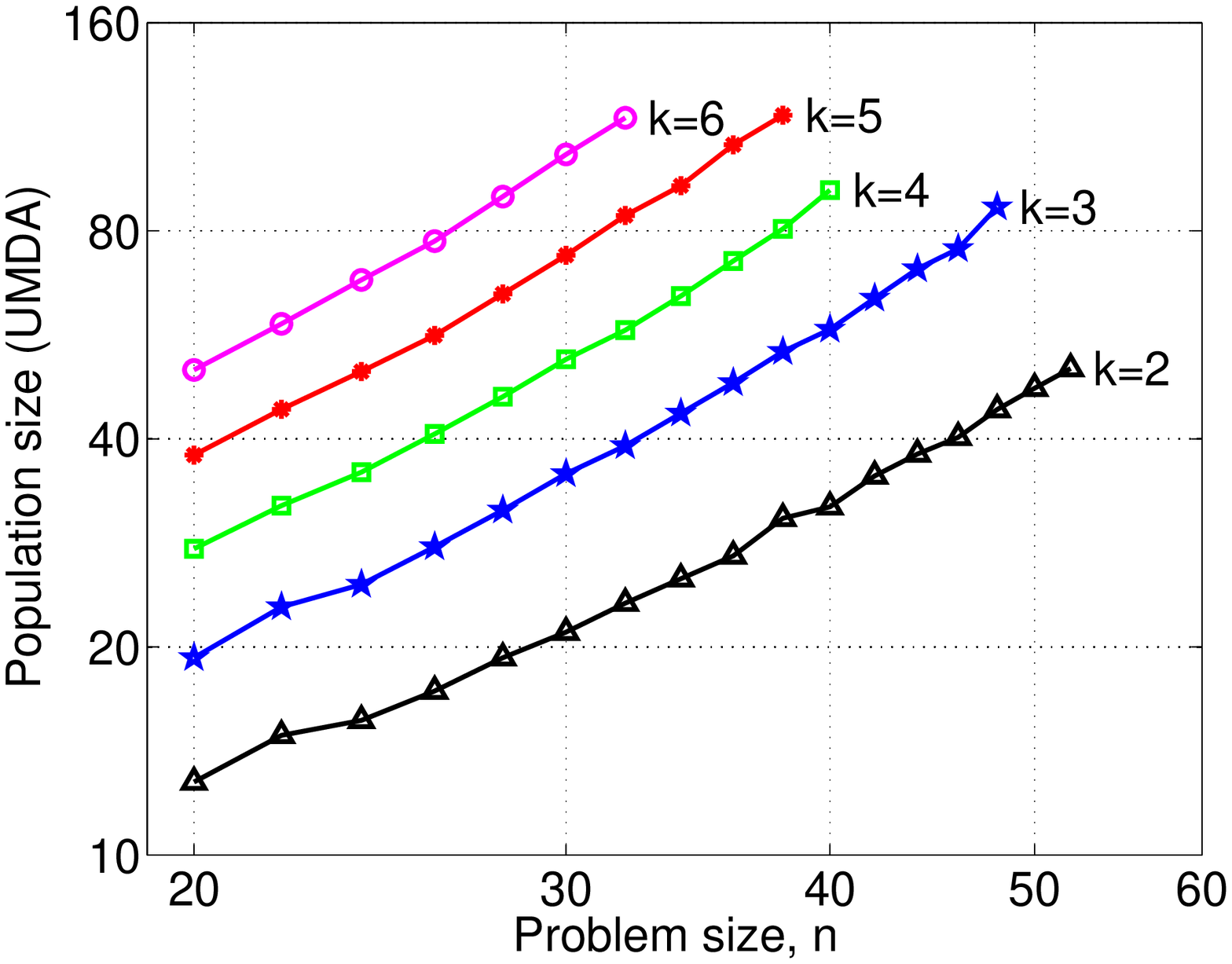,width=0.450\textwidth}}
\hspace*{3ex}
{\epsfig{file=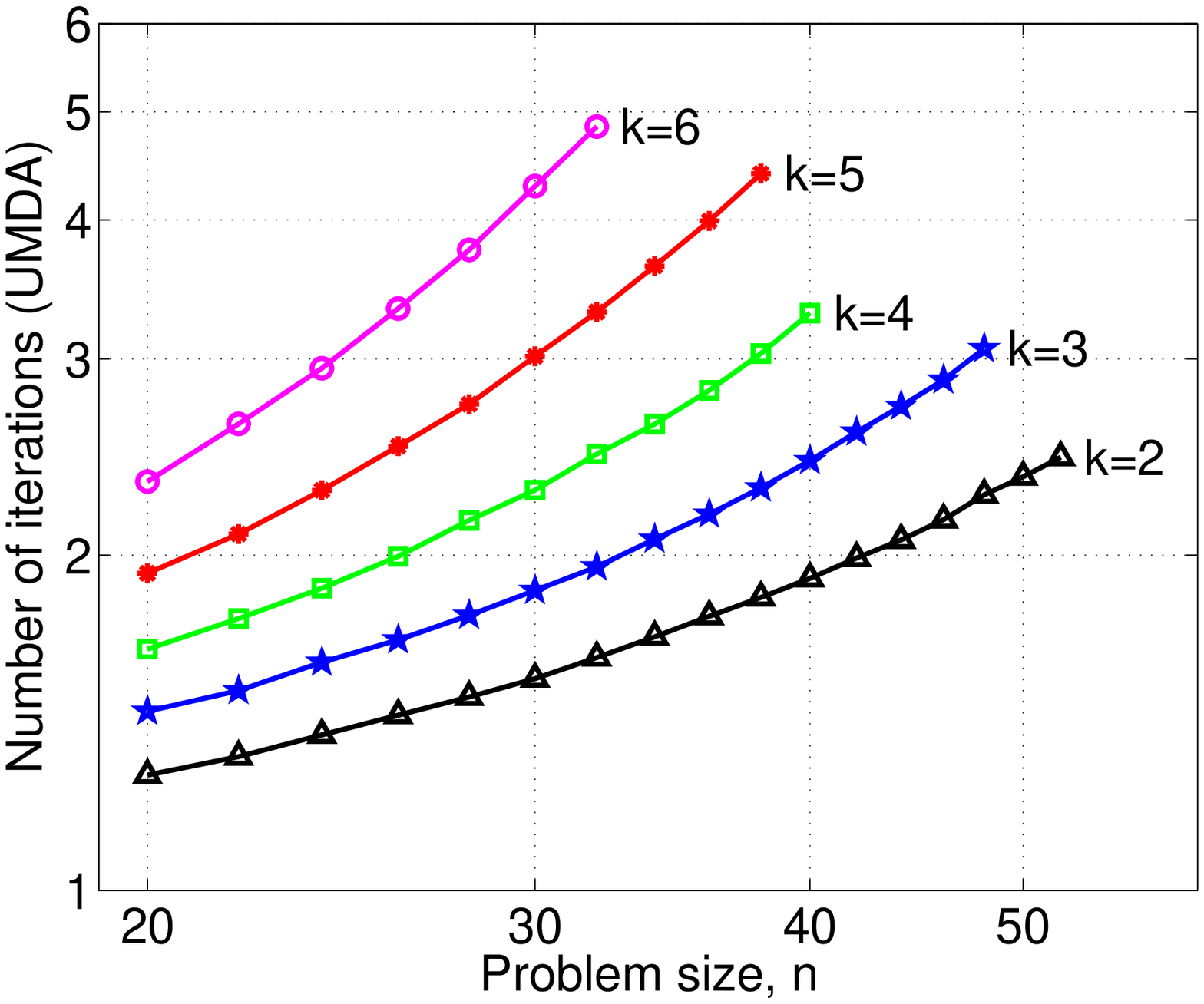,width=0.450\textwidth}}
\end{center}
\caption{Performance of UMDA on NK landscapes with $k=2$ to $6$ (log-log scale).}
\label{fig-umda-results}
\end{figure}

Figure~\ref{fig-umda-results} shows the average performance statistics for UMDA. Similarly as with hBOA, time complexity of UMDA grows exponentially fast with $k$ and its growth with $n$ for a fixed $k$ appears to be slightly faster than polynomial. 

\begin{figure}
\begin{center}
{\epsfig{file=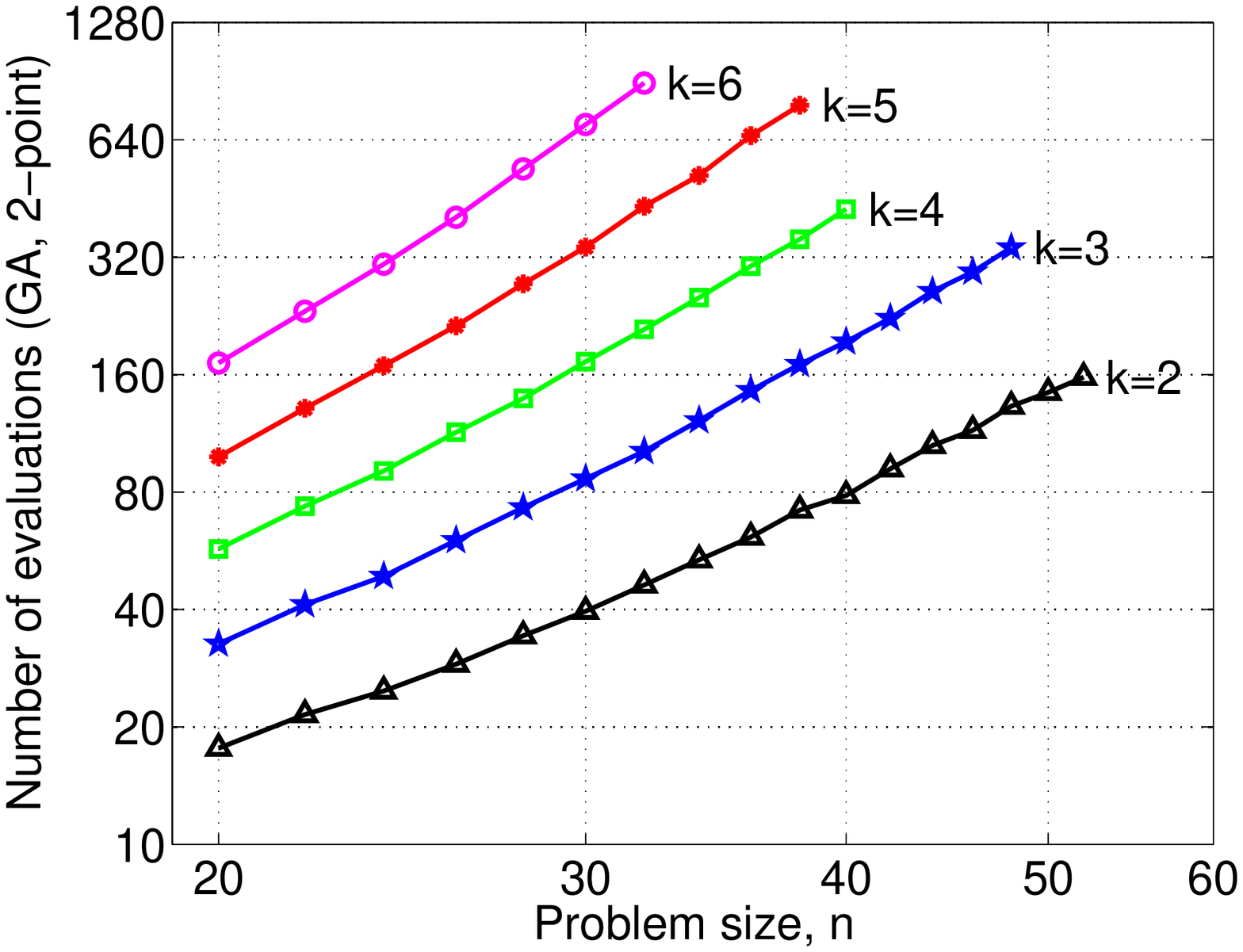,width=0.450\textwidth}}
\hspace*{3ex}
{\epsfig{file=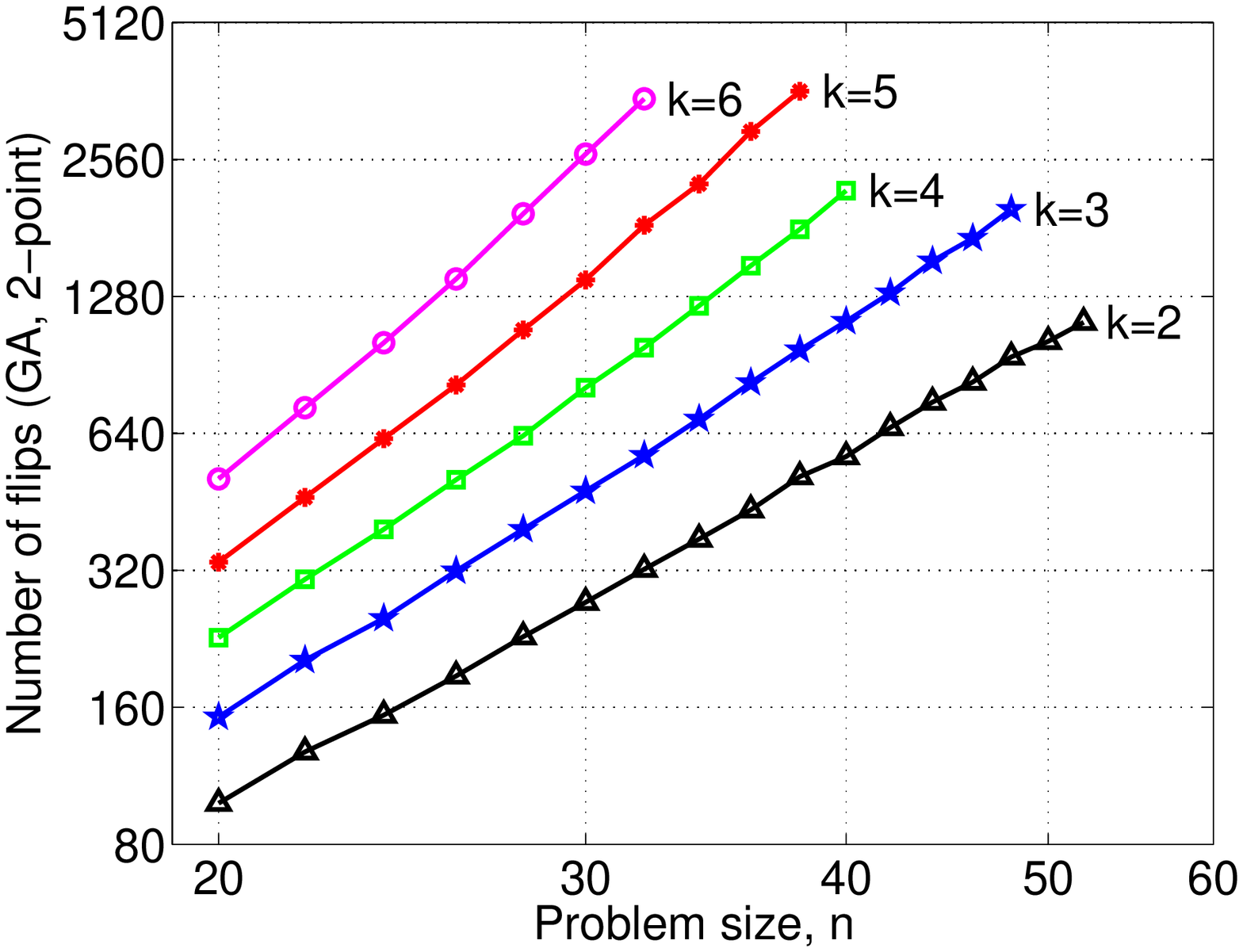,width=0.450\textwidth}}
{\epsfig{file=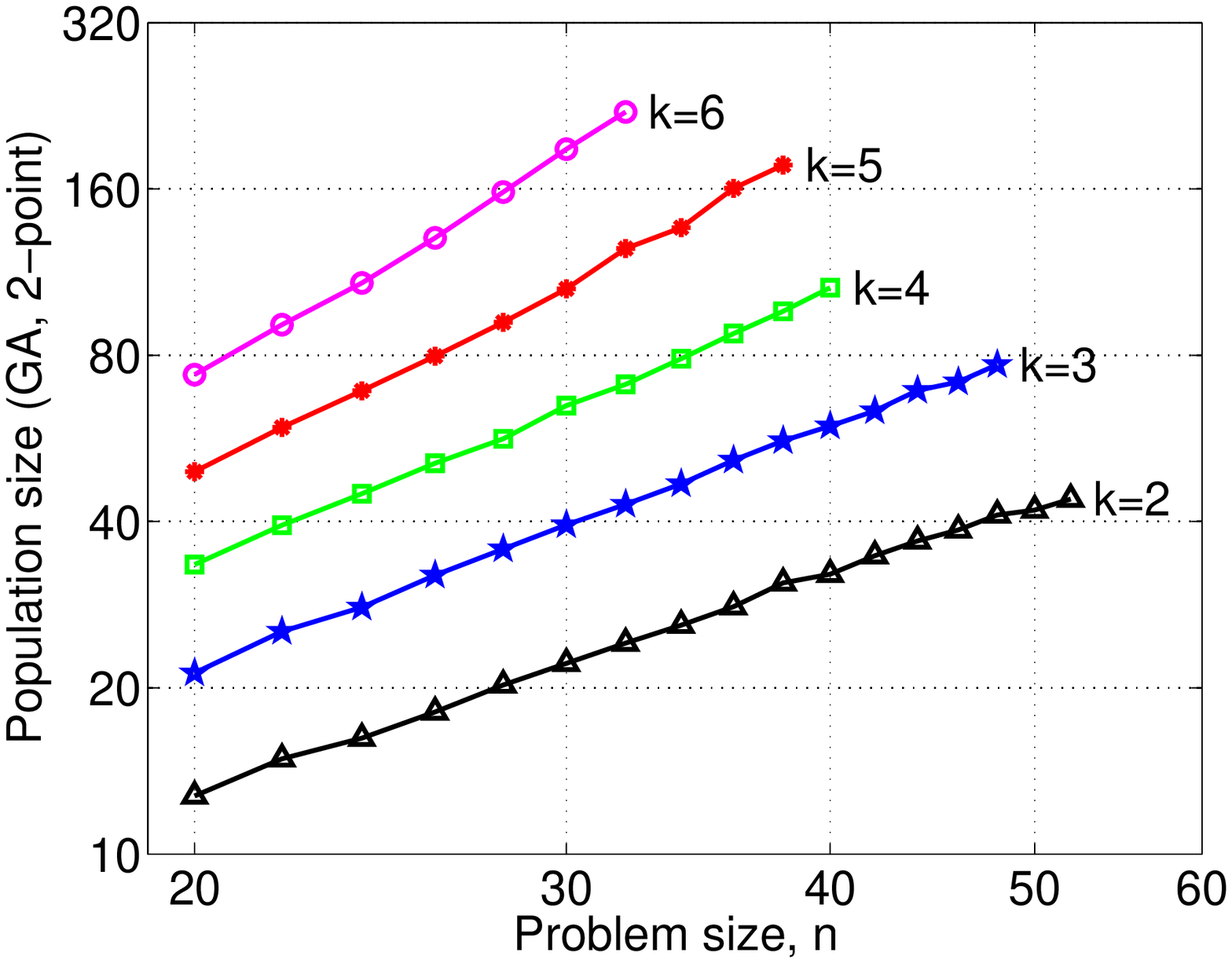,width=0.450\textwidth}}
\hspace*{3ex}
{\epsfig{file=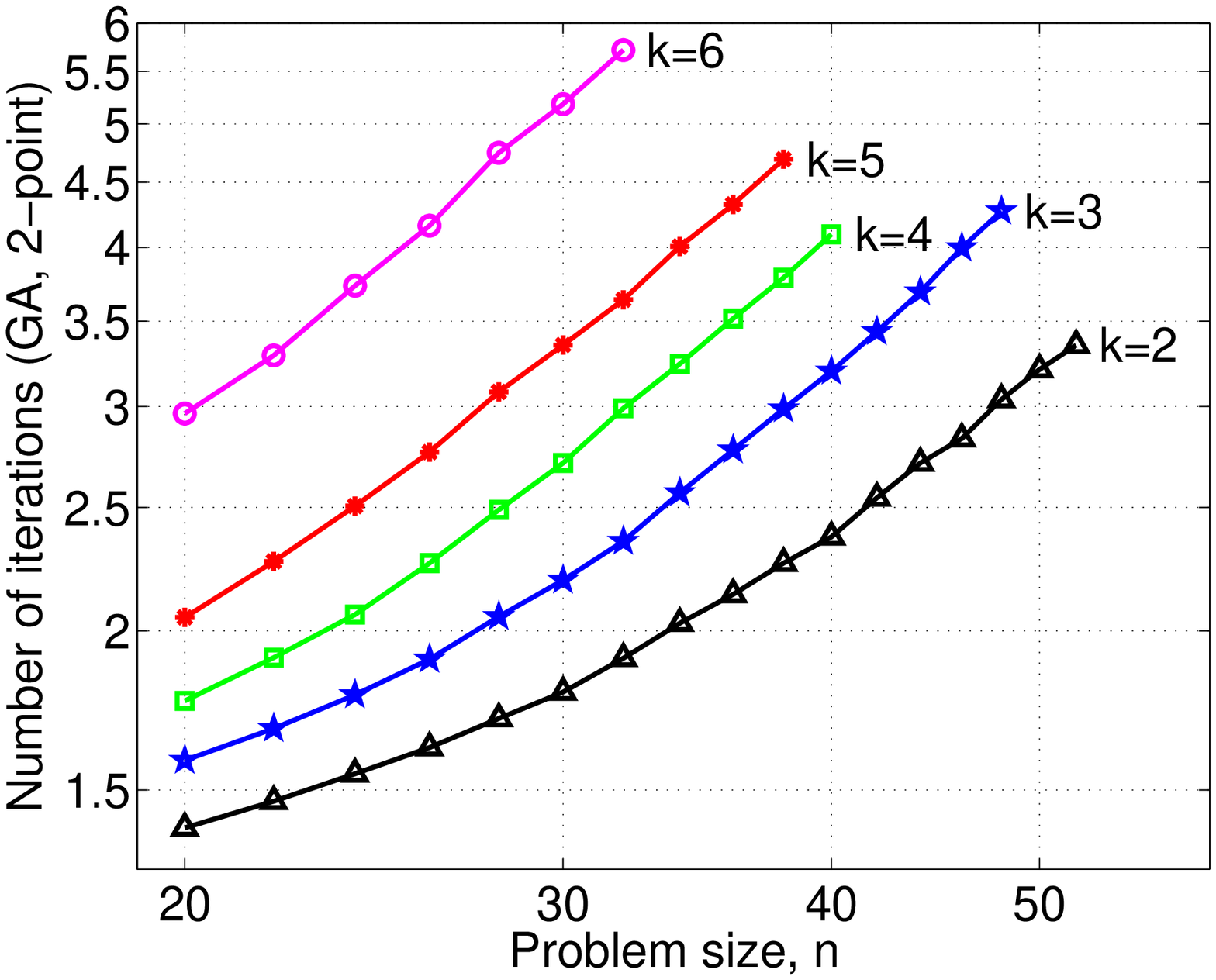,width=0.450\textwidth}}
\end{center}
\caption{Performance of GA with two-point crossover on NK landscapes with $k=2$ to $6$ (log-log scale).}
\label{fig-ga2p-results}
\end{figure}

\begin{figure}
\begin{center}
{\epsfig{file=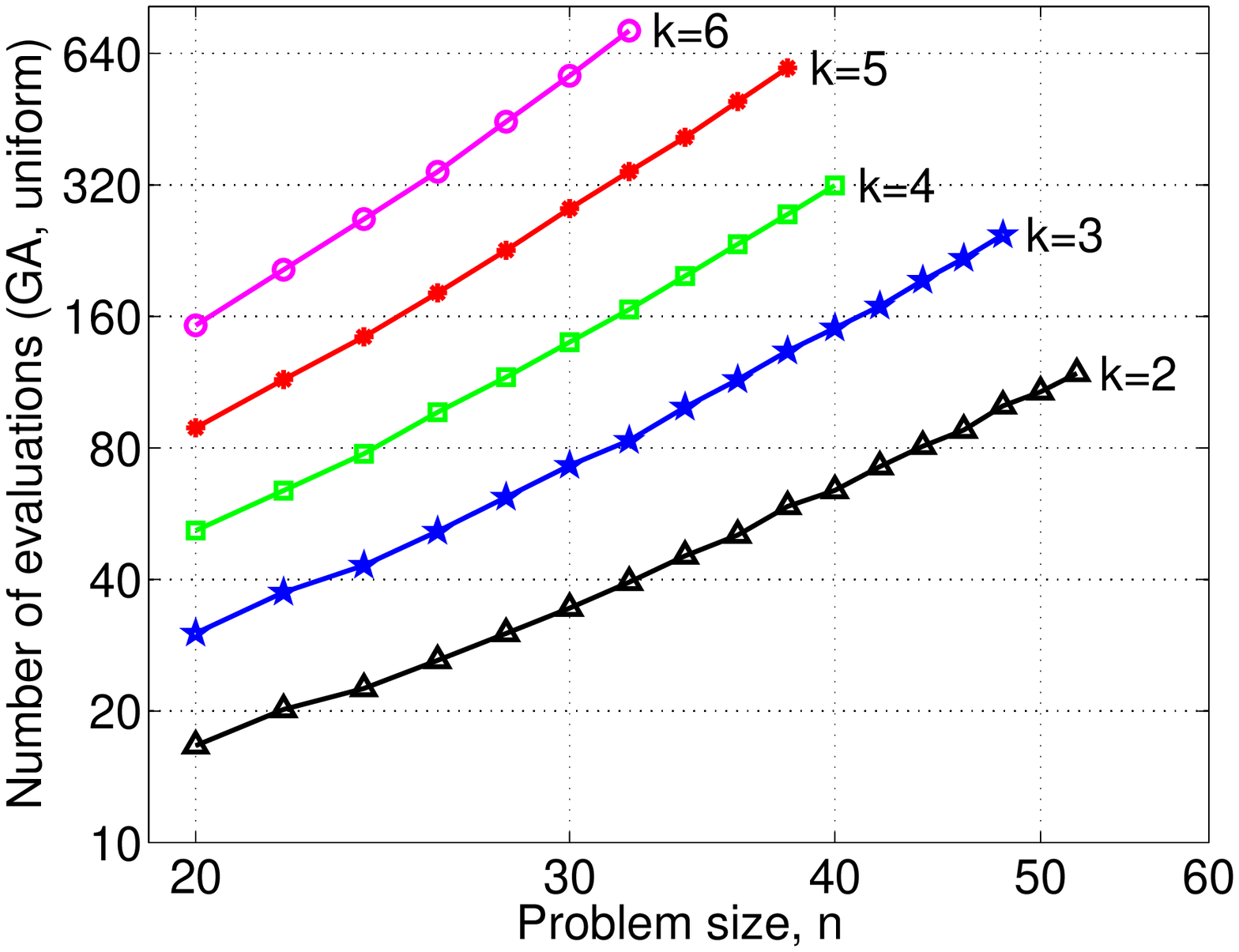,width=0.450\textwidth}}
\hspace*{3ex}
{\epsfig{file=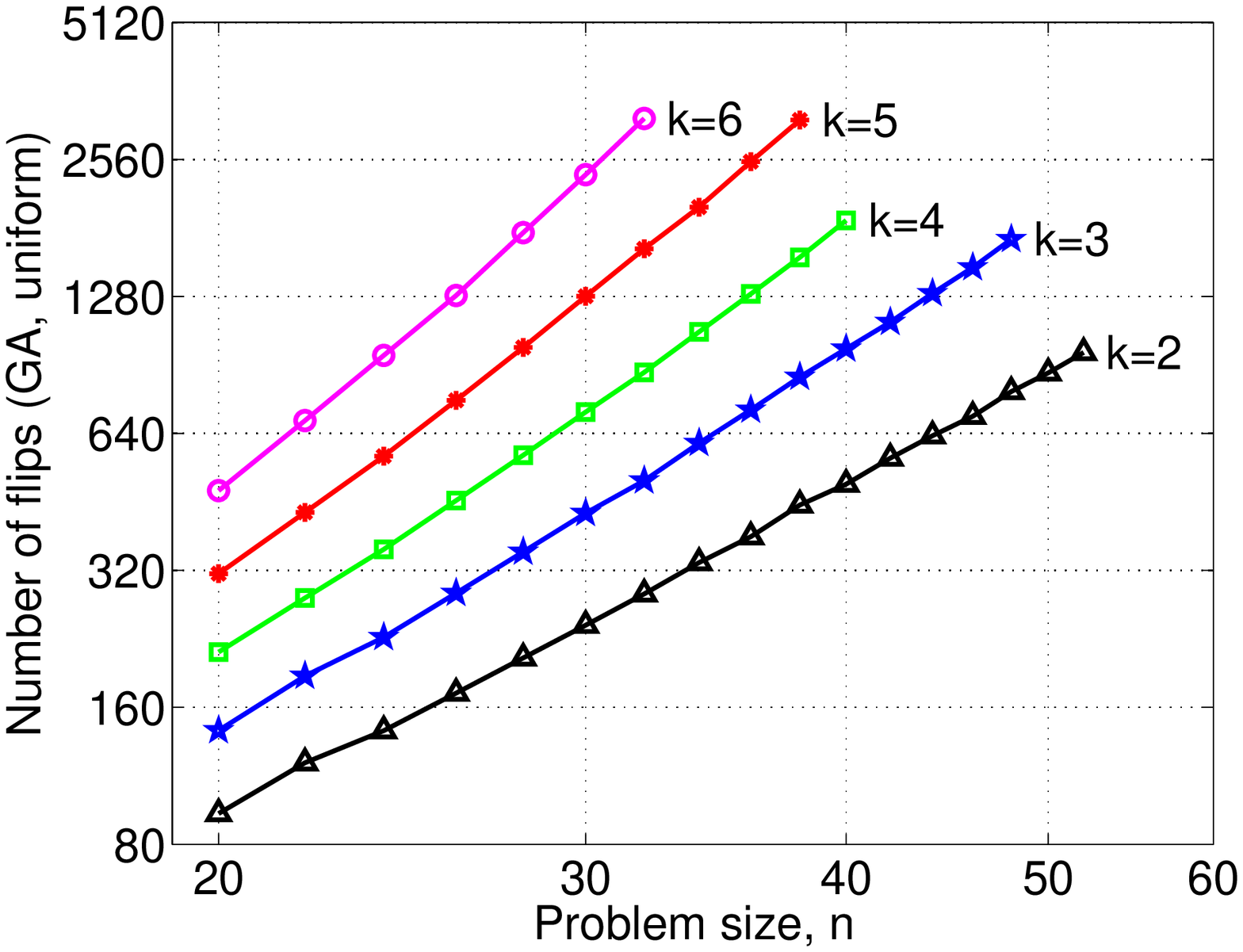,width=0.450\textwidth}}
{\epsfig{file=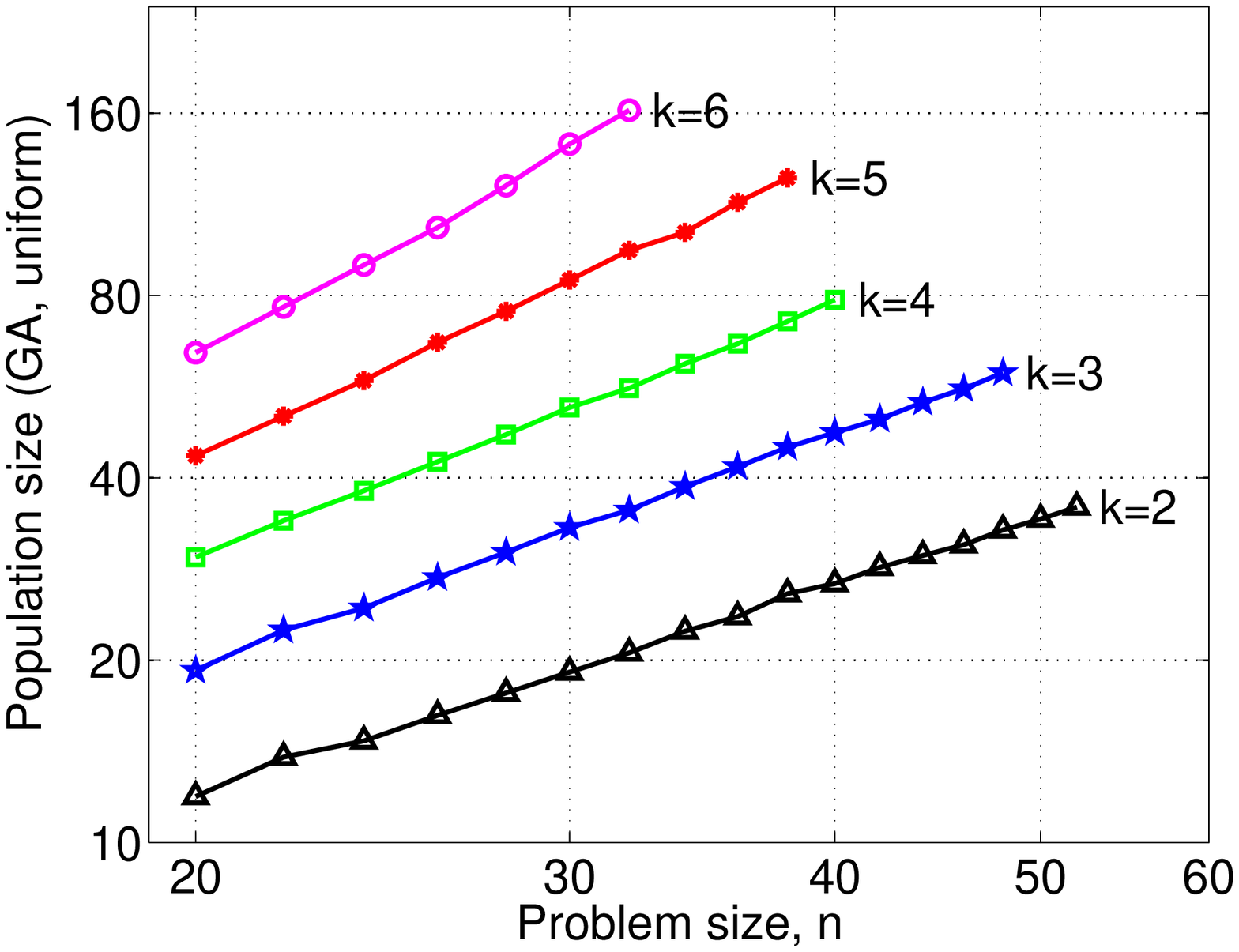,width=0.450\textwidth}}
\hspace*{3ex}
{\epsfig{file=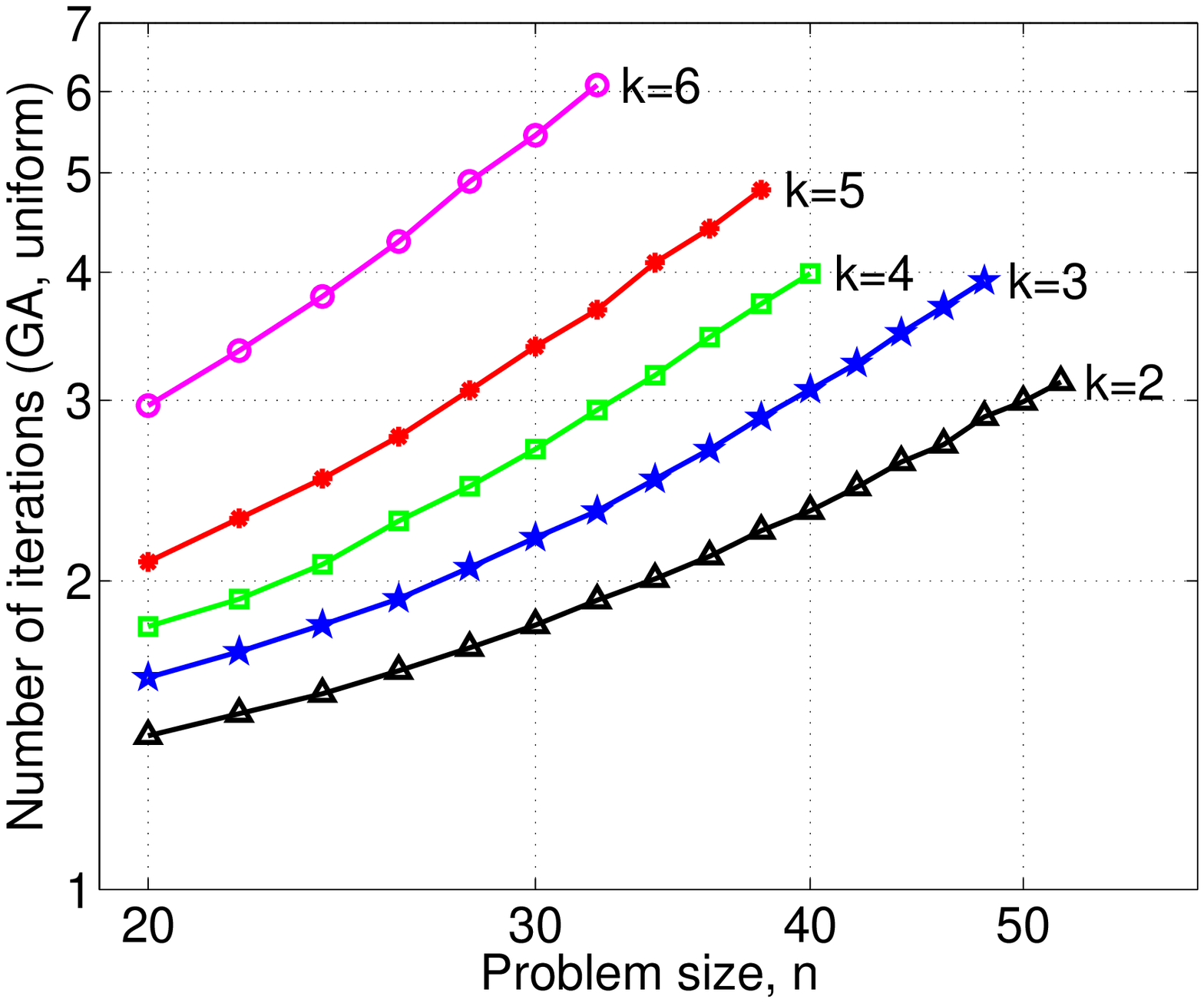,width=0.450\textwidth}}
\end{center}
\caption{Performance of GA with uniform crossover on NK landscapes with $k=2$ to $6$ (log-log scale).}
\label{fig-gau-results}
\end{figure}

\begin{figure}
\begin{center}
{\epsfig{file=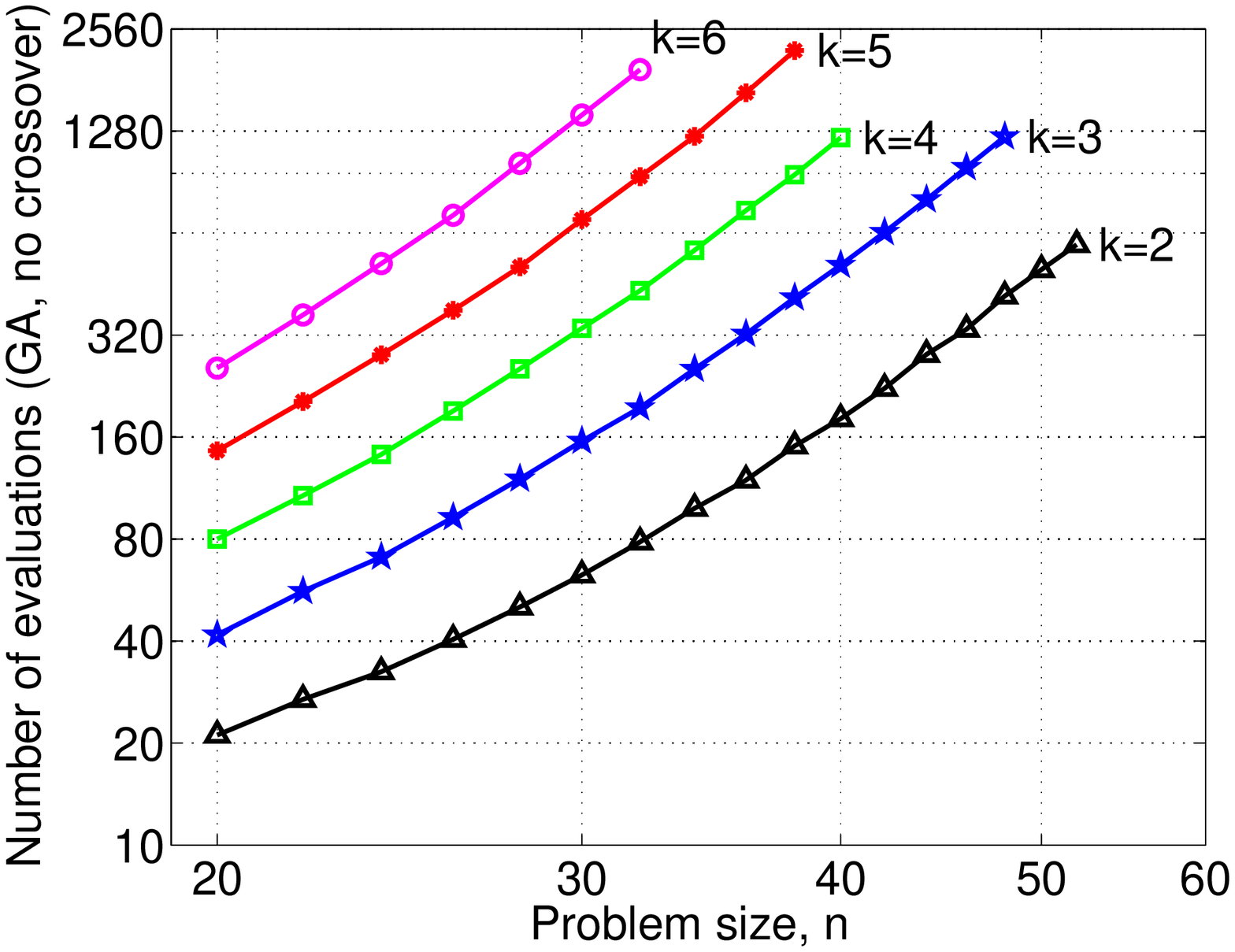,width=0.450\textwidth}}
\hspace*{3ex}
{\epsfig{file=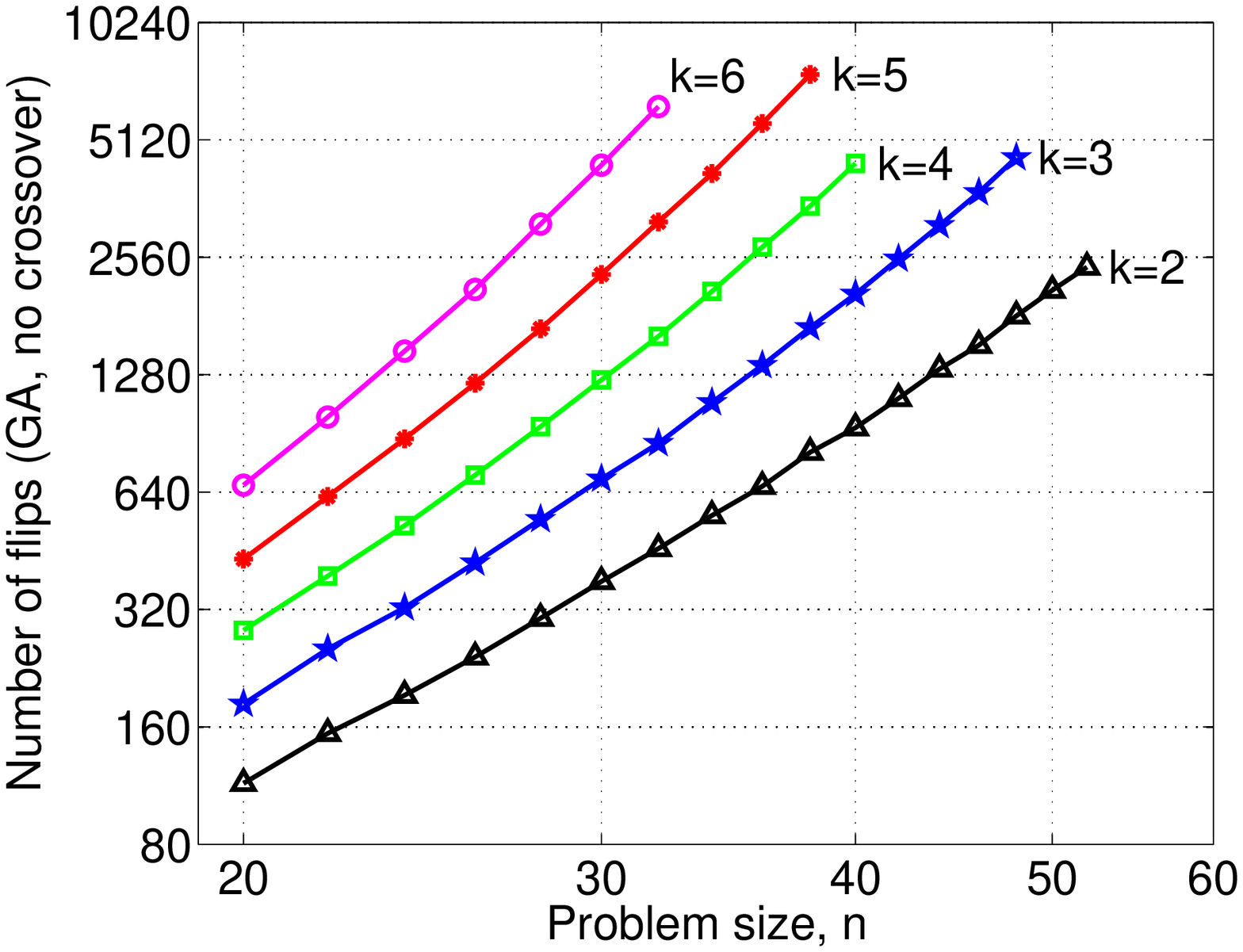,width=0.450\textwidth}}
{\epsfig{file=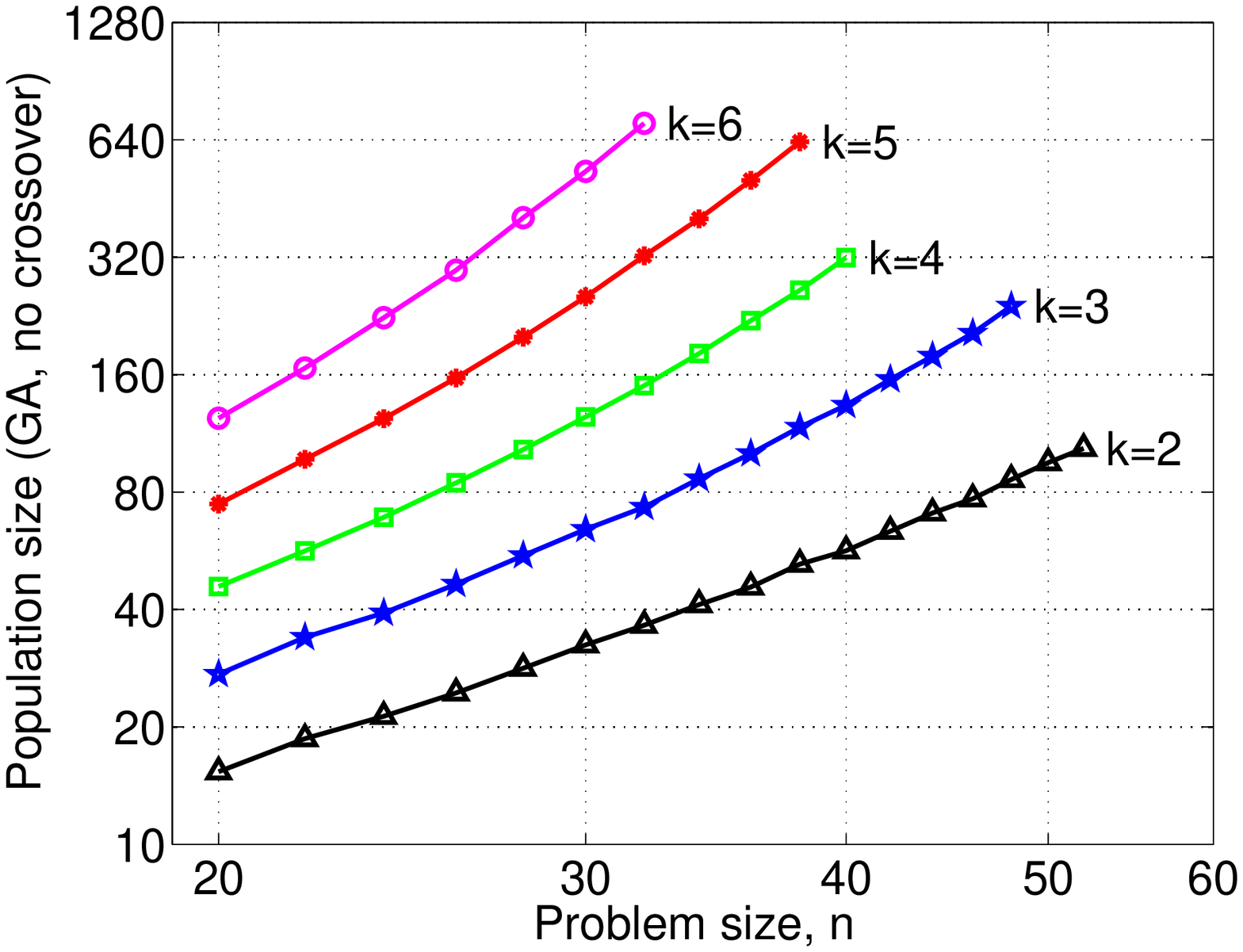,width=0.450\textwidth}}
\hspace*{3ex}
{\epsfig{file=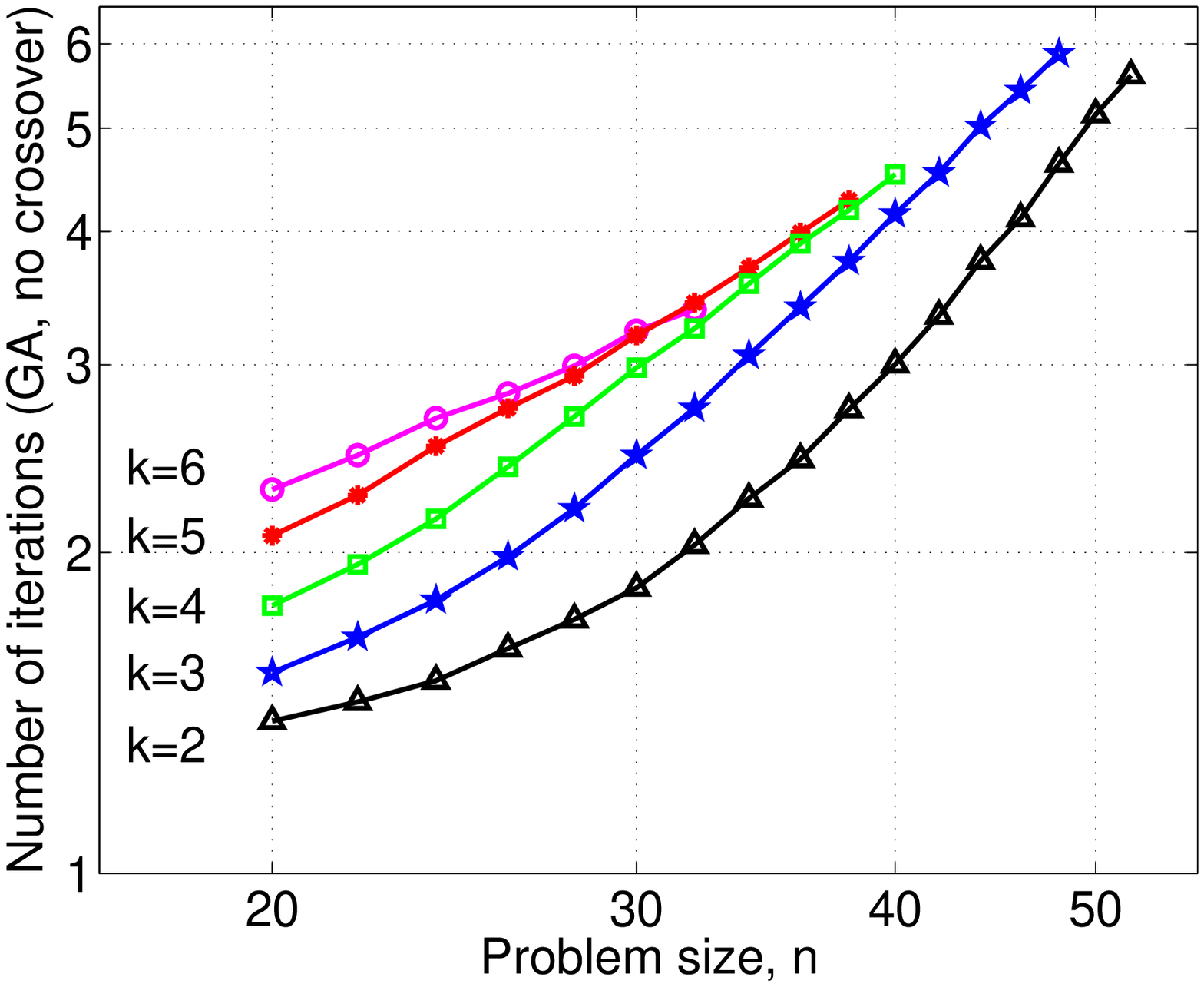,width=0.450\textwidth}}
\end{center}
\caption{Performance of GA with bit-flip mutation but without crossover on NK landscapes with $k=2$ to $6$ (log-log scale).}
\label{fig-ganc-results}
\end{figure}

Figures~\ref{fig-ga2p-results}, \ref{fig-gau-results} and~\ref{fig-ganc-results} show the average performance statistics for all three GA variants. Similarly as with hBOA and UMDA, time complexity of all GA variants grows exponentially fast with $k$ and its growth with $n$ for a fixed $k$ is slightly faster than polynomial. 

\subsection{Comparison of All Algorithms}
To compare performance of algorithms $A$ and $B$, for each problem instance, we can compute the ratio of the number of evaluations required by $A$ and the number of evaluations required by $B$; analogically, we can compute the ratio of the number of flips required by $A$ and the number of flips required by $B$. Then, the ratios can be averaged over all instances with specific $n$ and $k$. If $A$ performs better than $B$, the computed ratios should be smaller than $1$; if $A$ performs the same as $B$, the ratios should be about $1$; finally, if the $A$ performs worse than $B$, the ratios should be greater than $1$.

A comparison based on the aforementioned ratio was computed for each pair of algorithms studied in this work. To make the results easier to read, the superior algorithm was typically used as the second algorithm in the comparison (in the denominator of the ratios), so that the ratios should be expected to be greater than $1$. 

\begin{figure}
\centering
{\epsfig{file=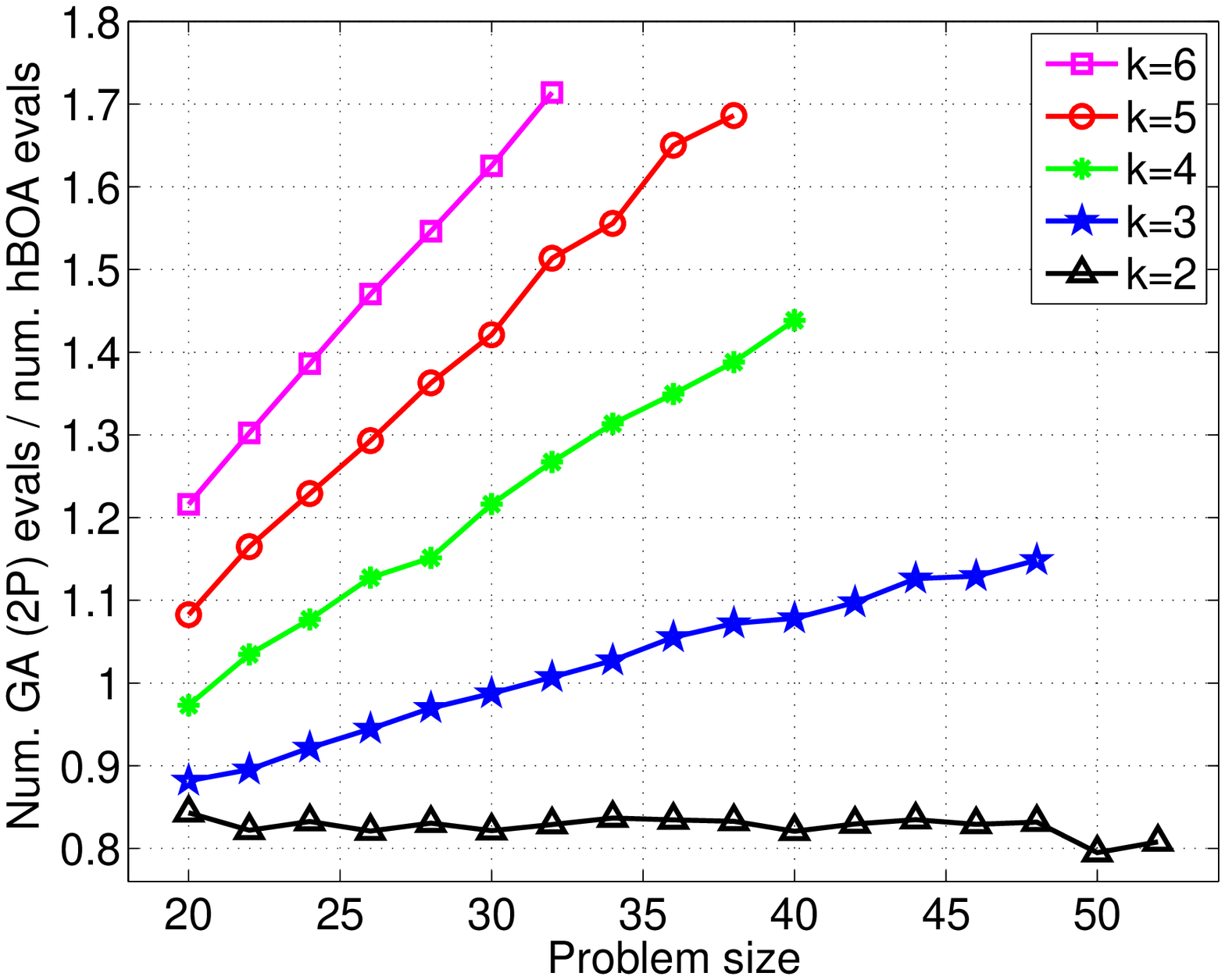,width=0.450\textwidth}}
\hspace*{3ex}
{\epsfig{file=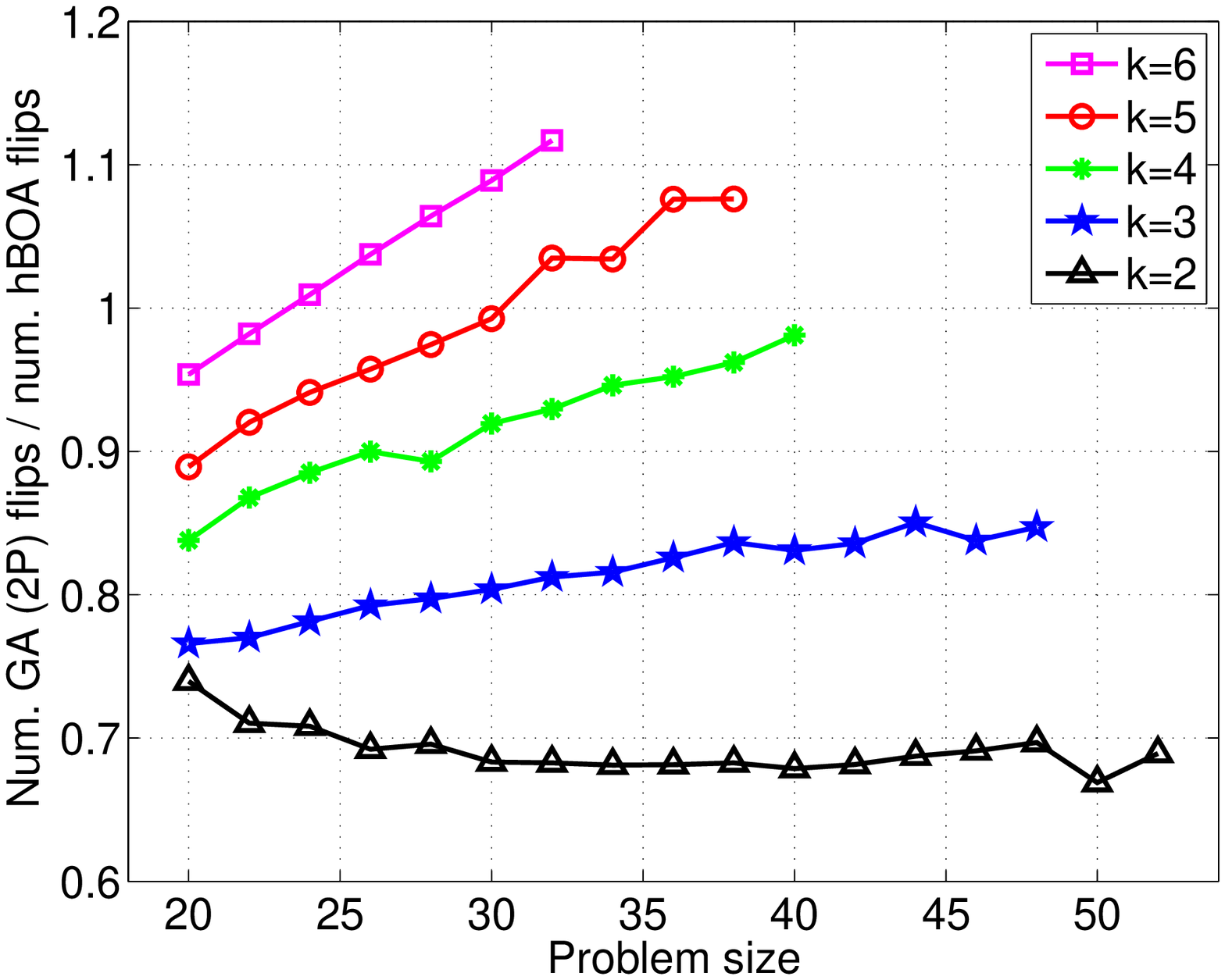,width=0.450\textwidth}}
\caption{Comparison of hBOA and GA with two-point crossover with respect to the number of evaluations and the number of flips. The comparison is visualized by the average ratio of the number of evaluations (number of flips) required by GA with two-point crossover and the number of evaluations (number of flips) required by hBOA. The greater the ratio, the better the performance of hBOA compared to GA. }
\label{fig-compare-hboa-ga2p}
\end{figure}

\begin{figure}
\centering
{\epsfig{file=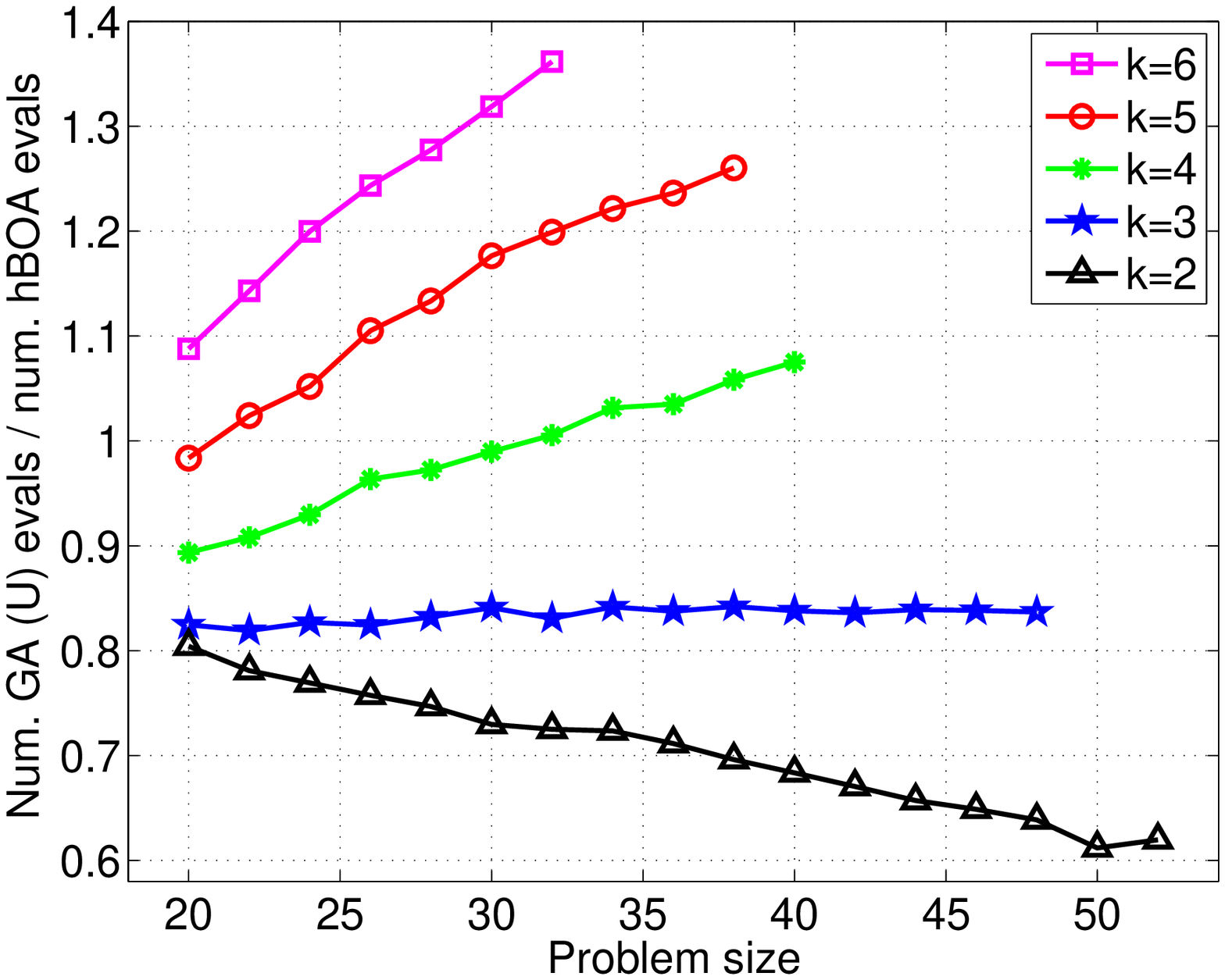,width=0.450\textwidth}}
\hspace*{3ex}
{\epsfig{file=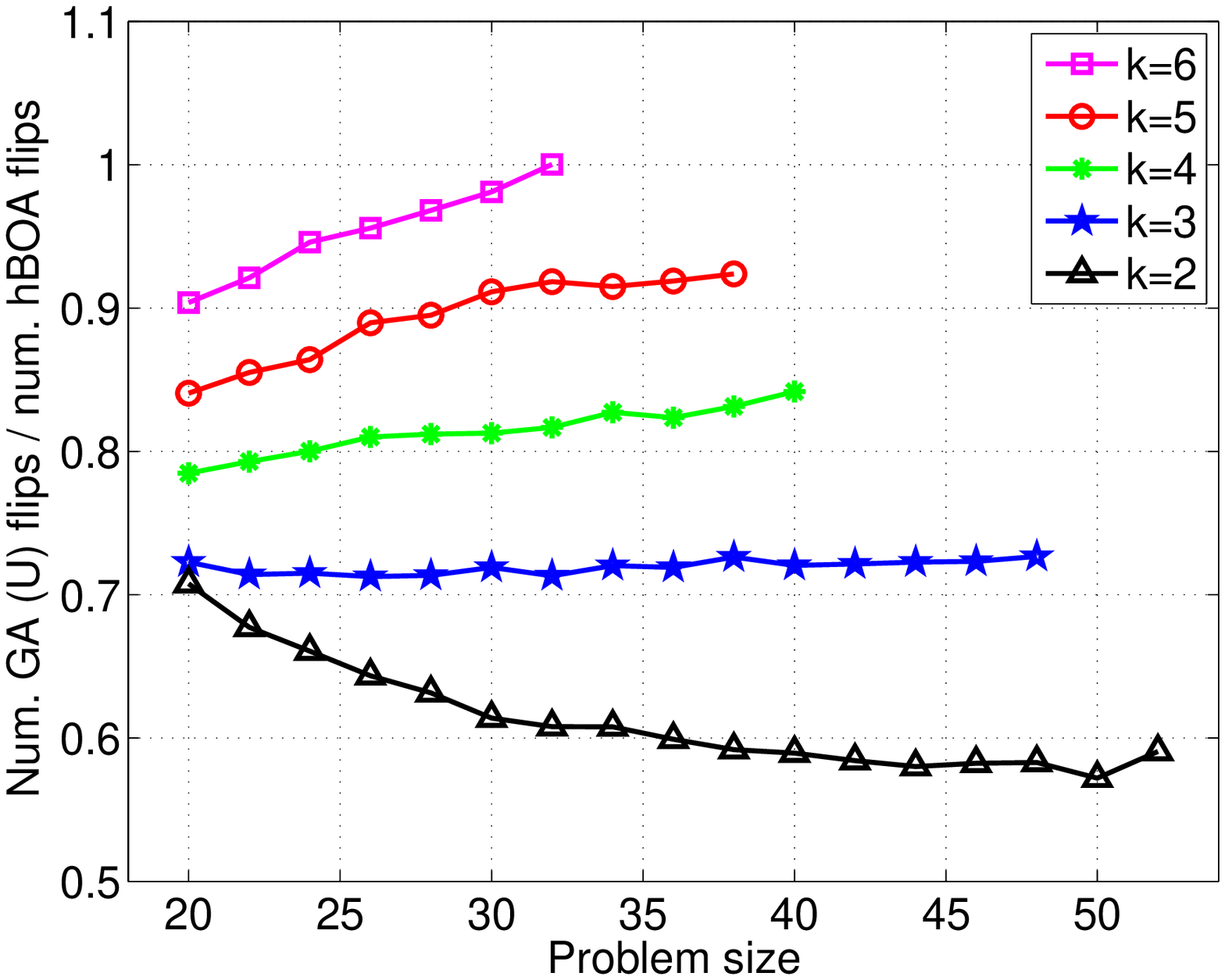,width=0.450\textwidth}}
\caption{Comparison of hBOA and GA with uniform crossover with respect to the number of evaluations and the number of flips. The comparison is visualized by the average ratio of the number of evaluations (number of flips) required by GA with uniform crossover and the number of evaluations (number of flips) required by hBOA. The greater the ratio, the better the performance of hBOA compared to GA.}
\label{fig-compare-hboa-gau}
\end{figure}

\begin{figure}
\centering
{\epsfig{file=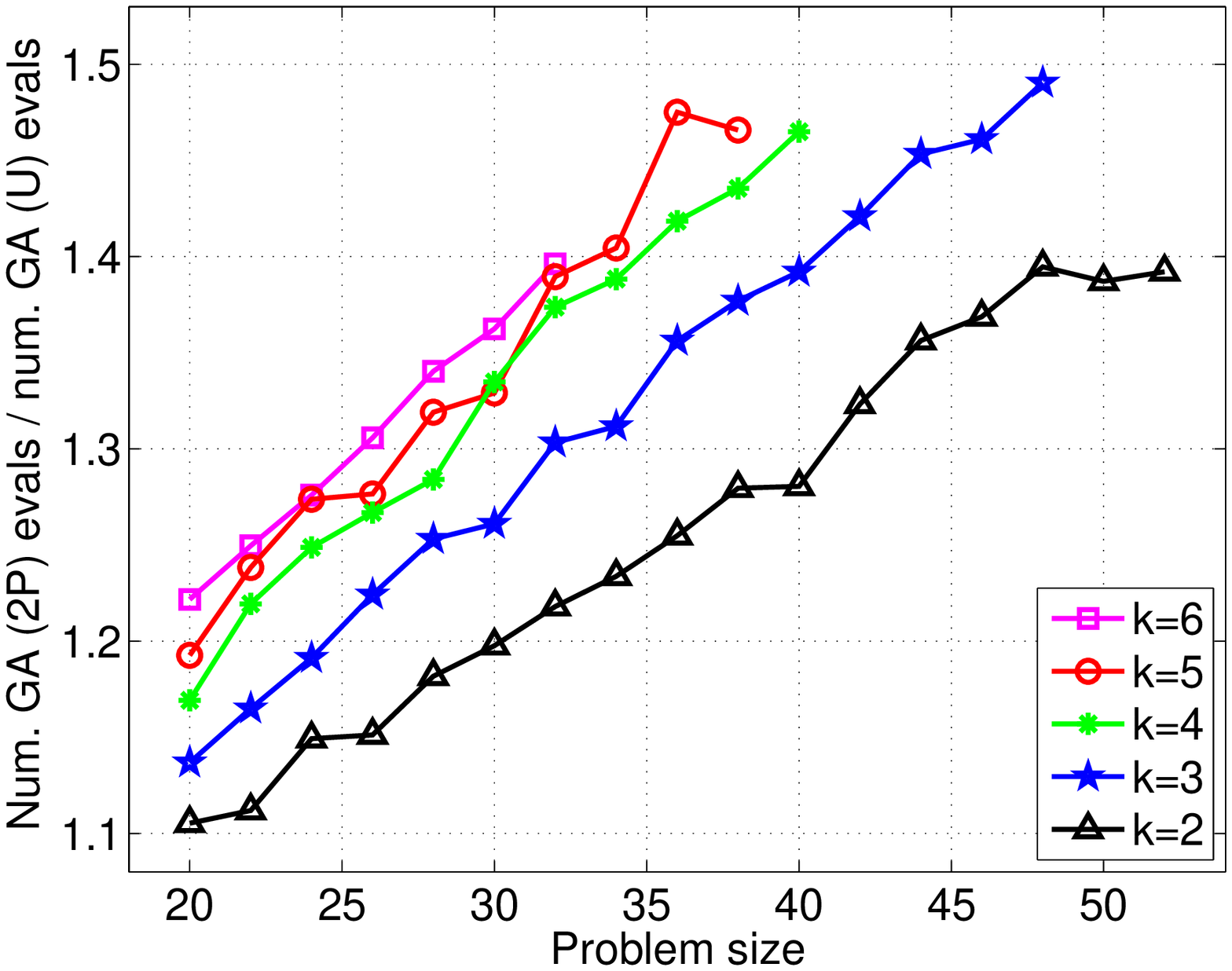,width=0.450\textwidth}}
\hspace*{3ex}
{\epsfig{file=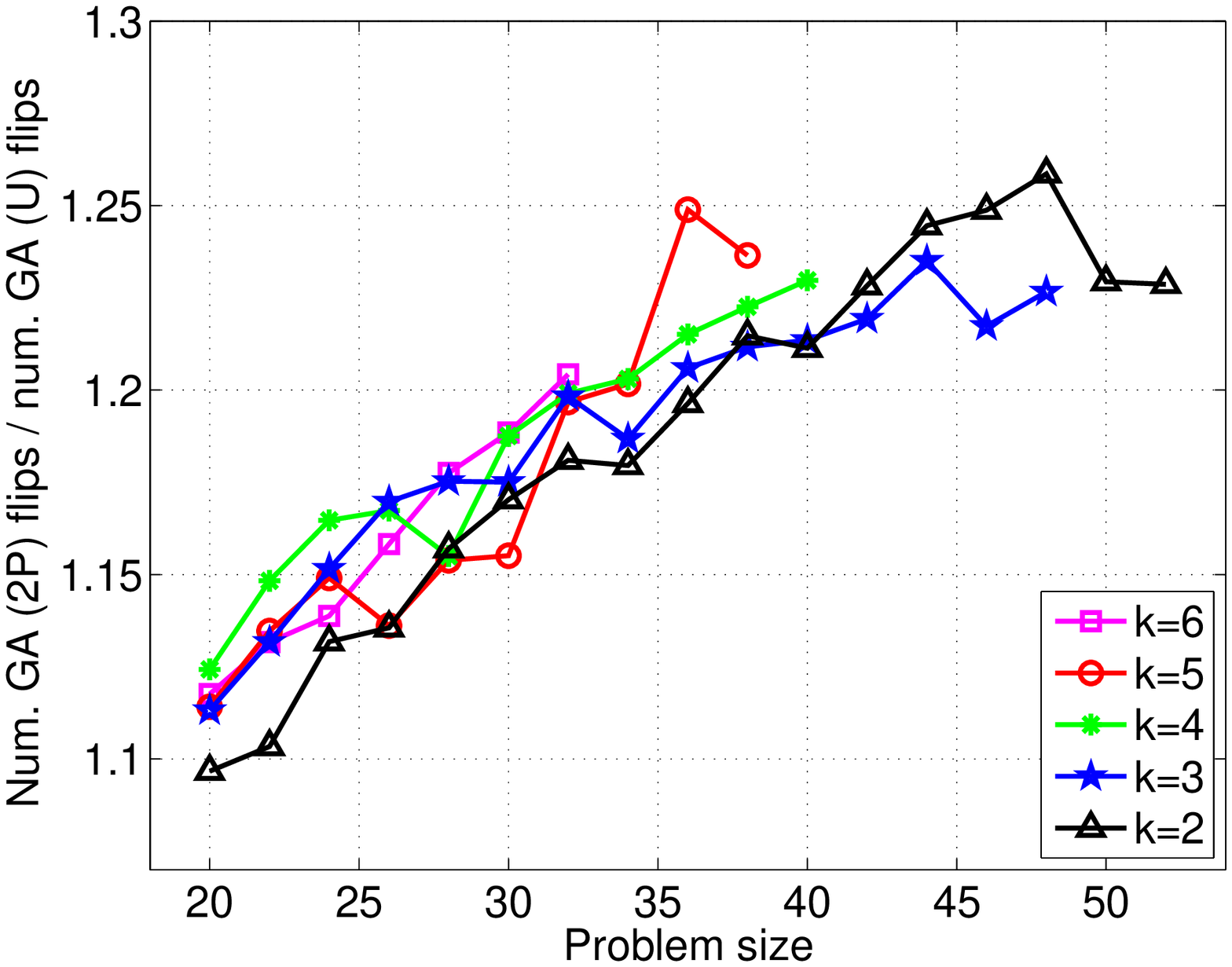,width=0.450\textwidth}}
\caption{Comparison of GA with uniform and two-point crossover with respect to the number of evaluations and the number of flips. The comparison is visualized by the average ratio of the number of evaluations (number of flips) required by GA with two-point crossover and the number of evaluations (number of flips) required by GA with uniform crossover. The greater the ratio, the better the performance of uniform crossover compared to two-point crossover.}
\label{fig-compare-gau-ga2p}
\end{figure}

\begin{figure}
\centering
{\epsfig{file=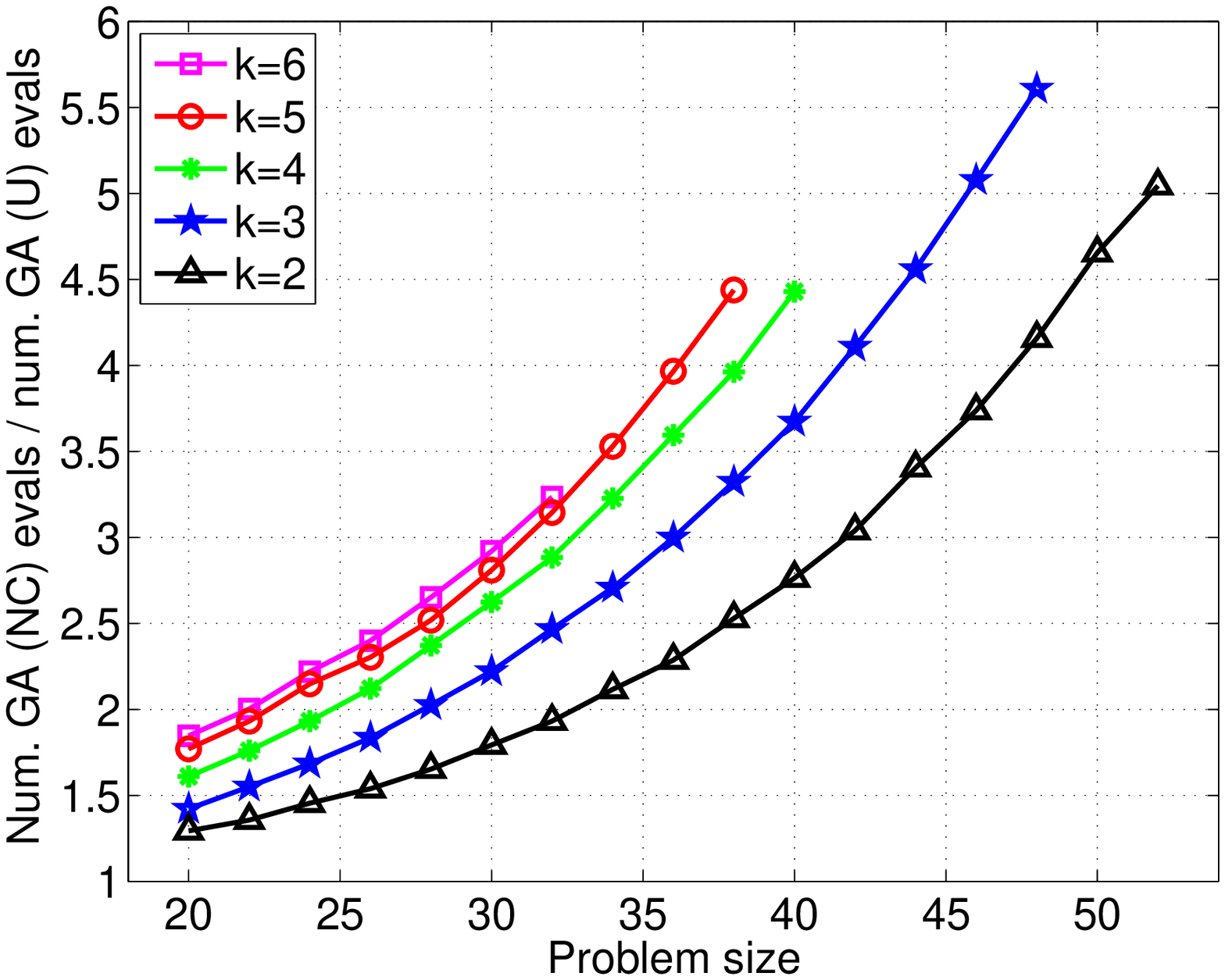,width=0.450\textwidth}}
\hspace*{3ex}
{\epsfig{file=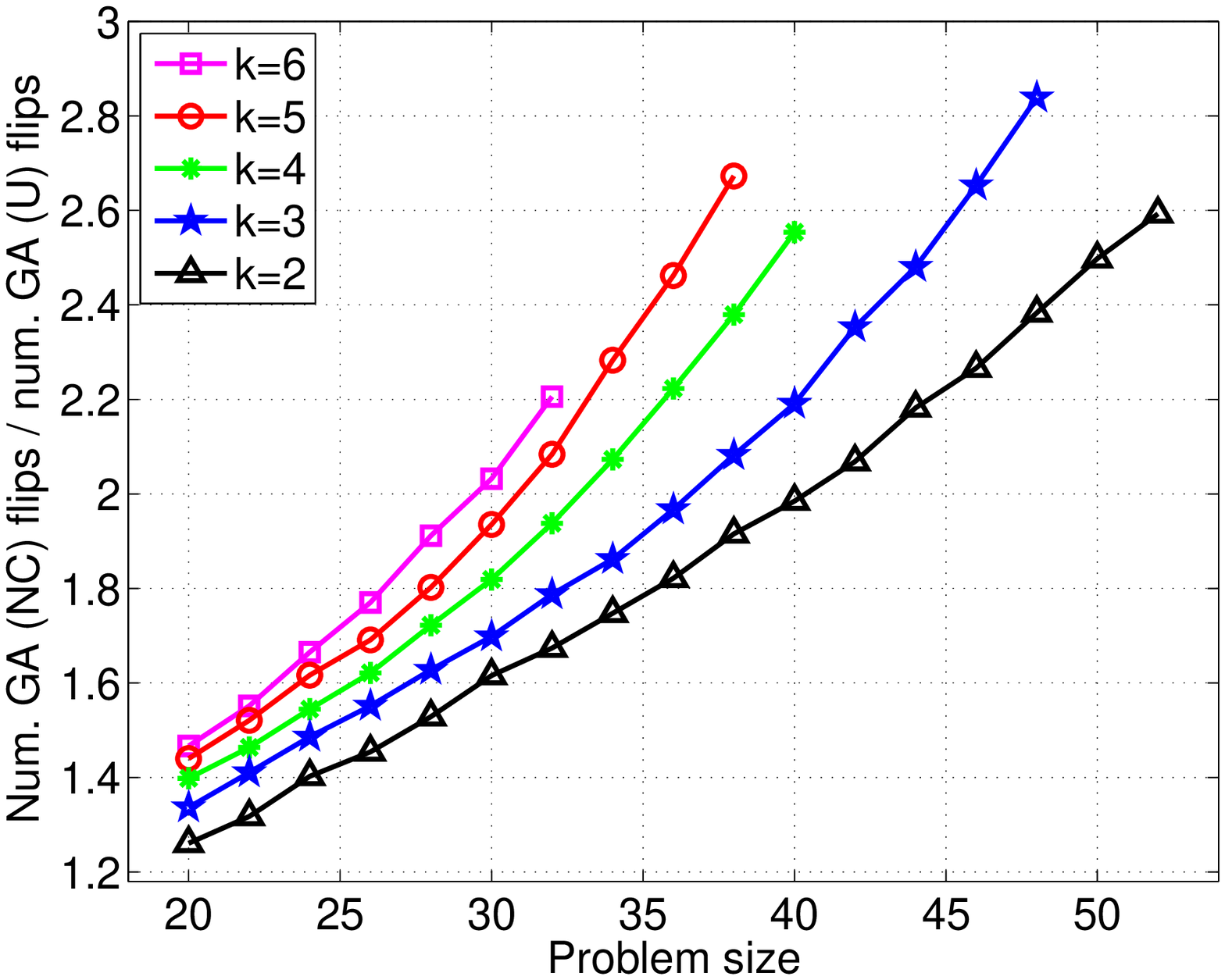,width=0.450\textwidth}}
\caption{Comparison of GA with uniform crossover and GA with no crossover (mutation only) with respect to the number of evaluations and the number of flips. The comparison is visualized by the average ratio of the number of evaluations (number of flips) required by GA without crossover and the number of evaluations (number of flips) required by GA with uniform crossover. The greater the ratio, the better the performance of uniform crossover compared to mutation only.}
\label{fig-compare-gau-ganc}
\end{figure}

\begin{figure}
\centering
{\epsfig{file=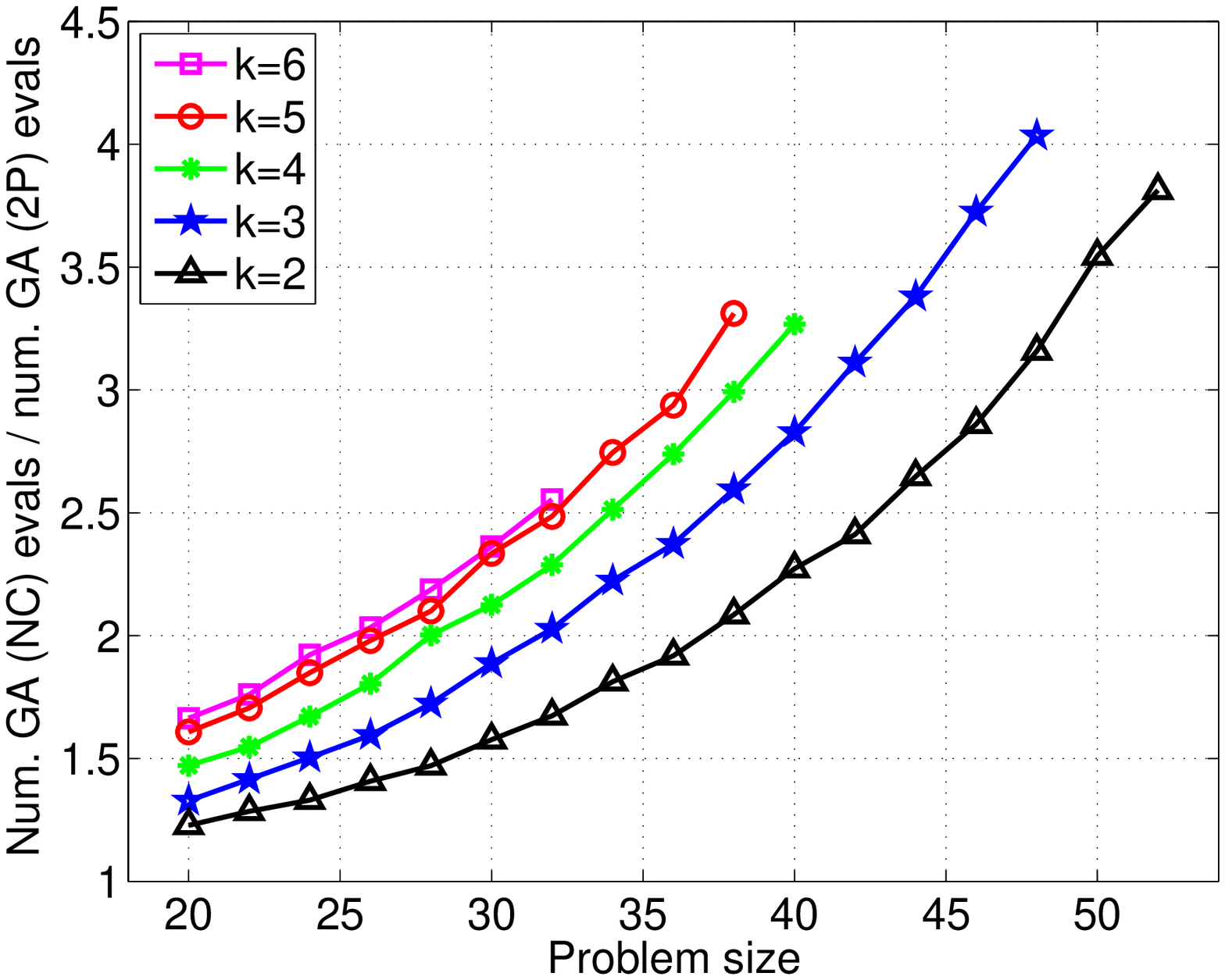,width=0.450\textwidth}}
\hspace*{3ex}
{\epsfig{file=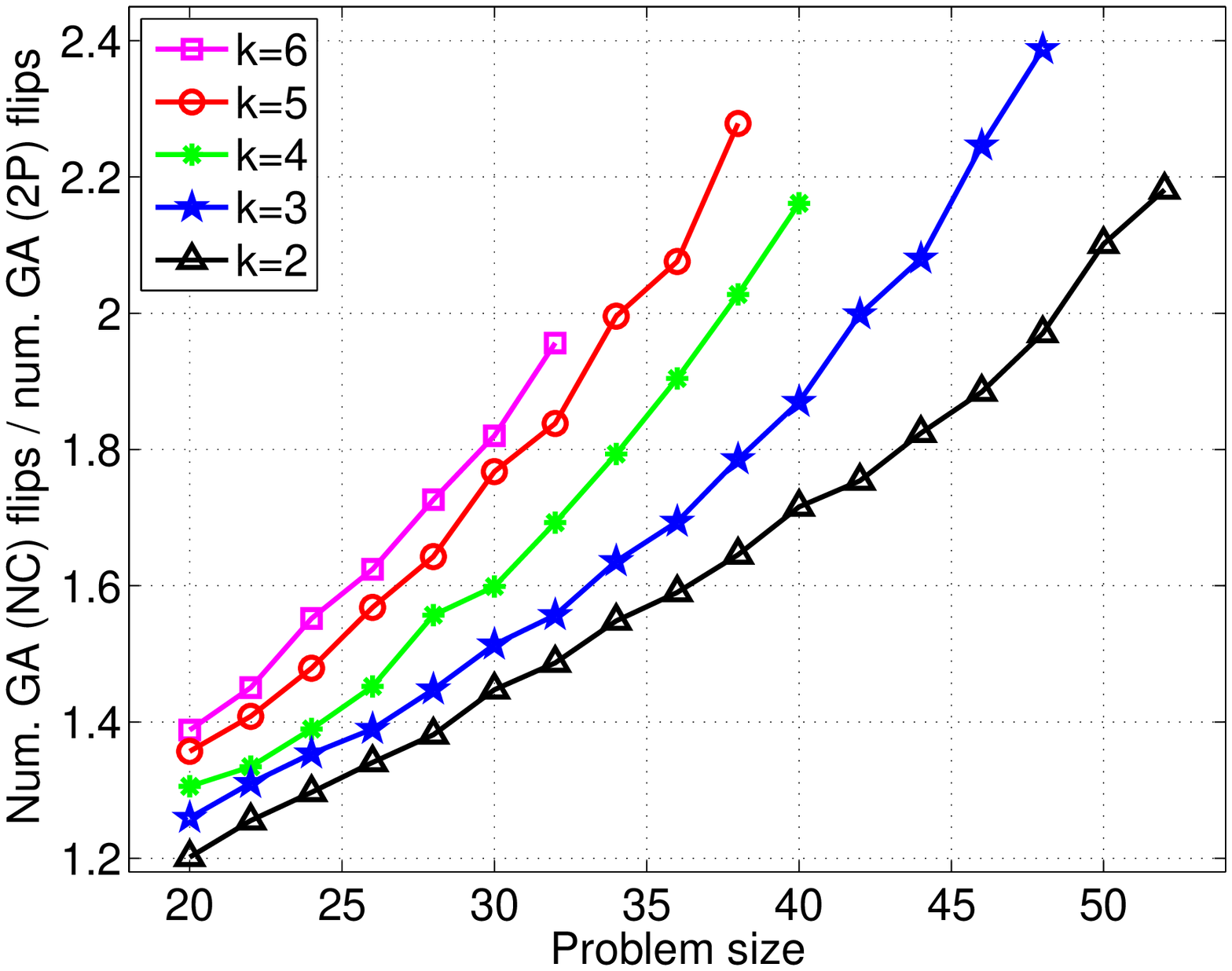,width=0.450\textwidth}}
\caption{Comparison of GA with two-point crossover and GA with no crossover (mutation only) with respect to the number of evaluations and the number of flips. The comparison is visualized by the average ratio of the number of evaluations (number of flips) required by GA without crossover and the number of evaluations (number of flips) required by GA with two-point crossover. The greater the ratio, the better the performance of two-point crossover compared to mutation only.}
\label{fig-compare-ga2p-ganc}
\end{figure}

Figure~\ref{fig-compare-hboa-ga2p} compares performance of GA with two-point crossover and that of hBOA. Figure~\ref{fig-compare-hboa-gau} compares performance of GA with uniform crossover and that of hBOA. Figure~\ref{fig-compare-gau-ga2p} compares performance of GA with two-point crossover and that of GA with uniform crossover. Finally, figures~\ref{fig-compare-gau-ganc} and~\ref{fig-compare-ga2p-ganc} compare performance of GAs with and without crossover. 

One of the important trends to observe in the results of the comparisons is the change in the two ratios with problem size. In most cases, when one algorithm outperforms another one, the differences become more significant as the problem size increases. In some cases, although one algorithm outperforms another one on small problems, because of the observed dynamics with problem size, we can expect the situation to reverse for large problems. 

The comparisons based on the number of evaluations and the number of flips can be summarized as follows:
\begin{description}
\item[hBOA.] While for small values of $k$, hBOA is outperformed by other algorithms included in the comparison, as $k$ increases, the situation changes rapidly. More specifically, for larger $k$, hBOA outperforms all other algorithms and its relative performance with respect to other algorithms improves with increasing problem size. The larger the $k$, the more favorably hBOA compares to other algorithms. 
\item[GA with uniform crossover.] GA with uniform crossover performs better than GA with two-point crossover and UMDA regardless of $k$ and its relative performance with respect to these algorithms improves with problem size. However, as mentioned above, for larger values of $k$, GA with uniform crossover is outperformed by hBOA and the factor by which hBOA outperforms GA with uniform crossover grows with problem size. 
\item[GA with two-point crossover.] GA with two-point crossover performs worse than hBOA and GA with uniform crossover for larger values of $k$, but it still outperforms UMDA with respect to the number of flips, which is the most important performance measure. 
\item[UMDA.] UMDA performs worst of all recombination-based algorithms included in the comparison except for a few cases with small values of $k$. 
\item[Crossover versus mutation.] Crossover has proven to outperform pure mutation, which is clear from all the results. First of all, for the most difficult instances, hBOA---which is a pure selectorecombinative evolutionary algorithm with no explicit mutation---outperforms other algorithms with increasing $n$. Second, eliminating crossover from GA significantly decreases its efficiency and the mutation-based approaches perform worst of all compared algorithms. Specifically, GA with no crossover is outperformed by all other variants of GA, and the stochastic hill climbing is not even capable of solving many problem instances in practical time. 
\end{description} 

\begin{figure}
\centering
{\epsfig{file=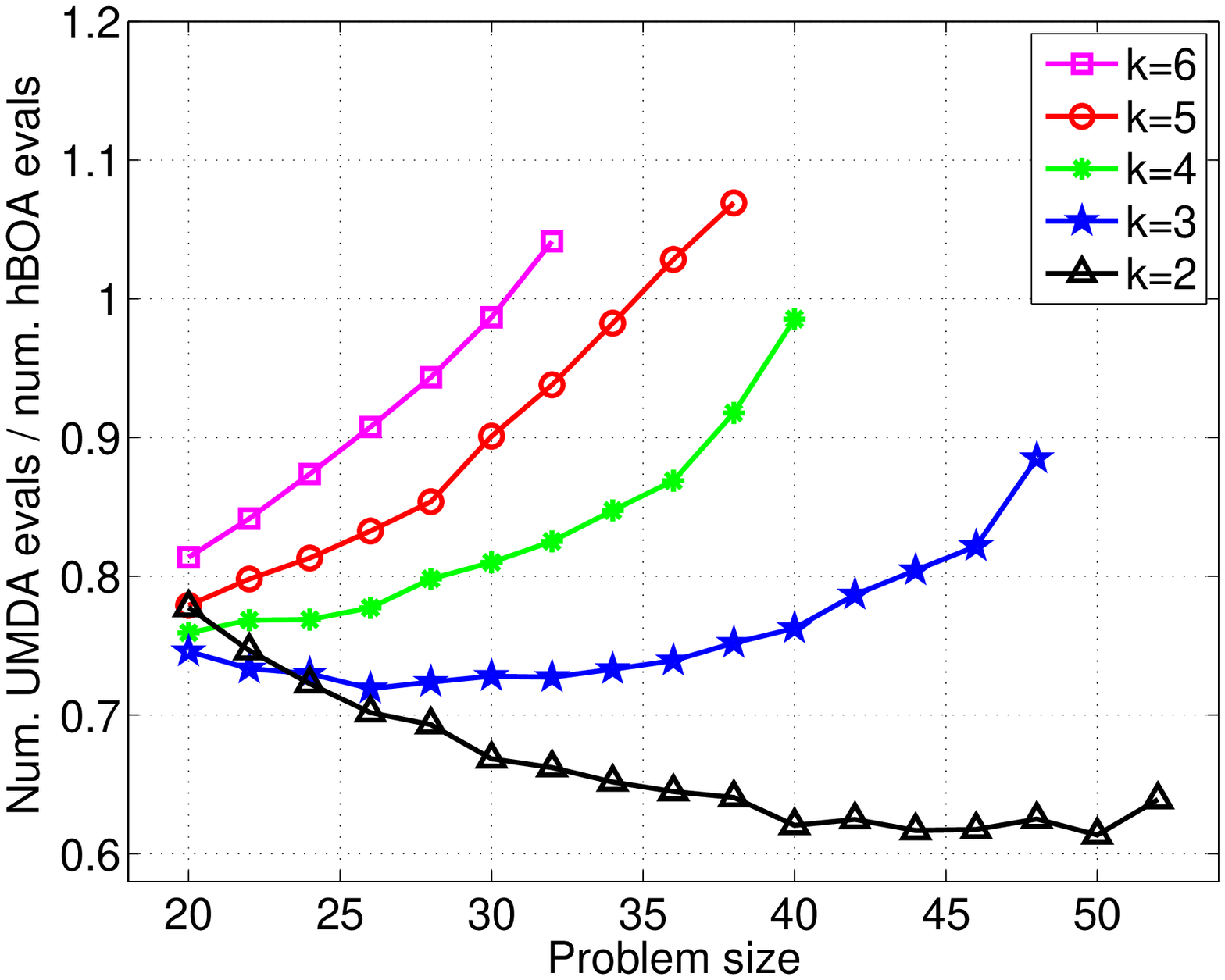,width=0.450\textwidth}}
\hspace*{3ex}
{\epsfig{file=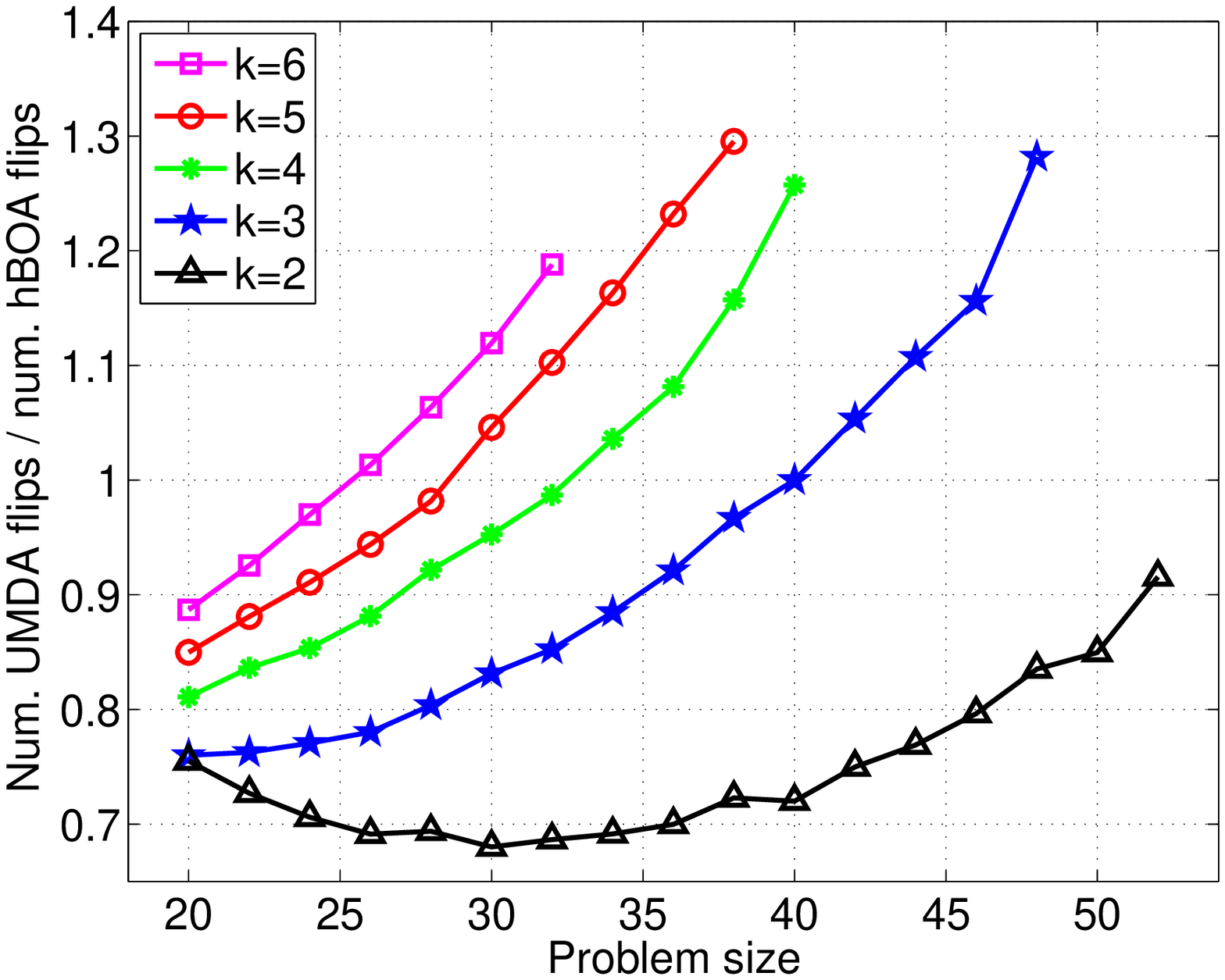,width=0.450\textwidth}}
\caption{Comparison of hBOA and UMDA with respect to the number of evaluations and the number of flips. The comparison is visualized by the average ratio of the number of evaluations (number of flips) required by UMDA and the number of evaluations (number of flips) required by hBOA. The greater the ratio, the better the performance of hBOA compared to UMDA.}
\label{fig-compare-hboa-umda}
\end{figure}

\begin{figure}
\centering
{\epsfig{file=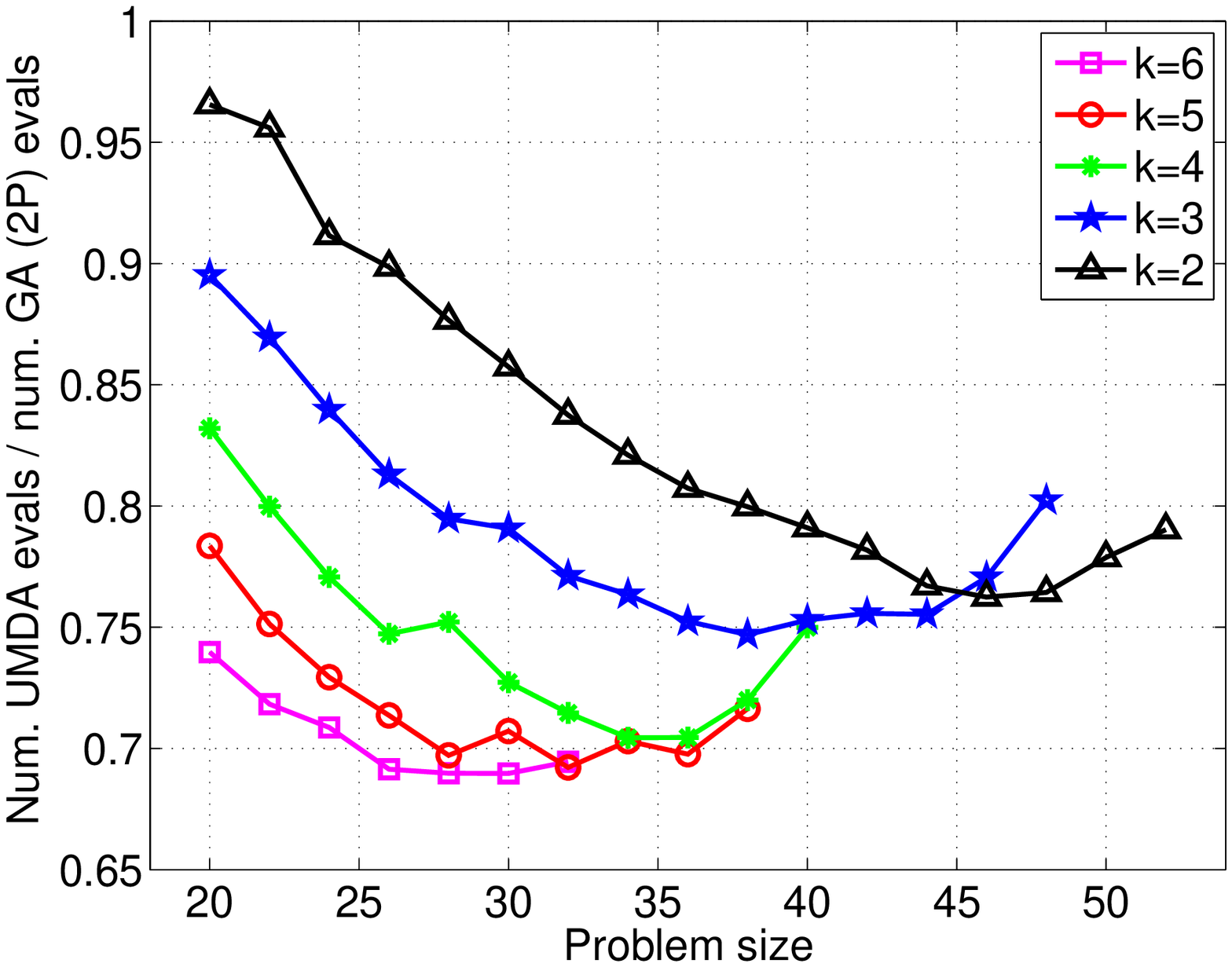,width=0.450\textwidth}}
\hspace*{3ex}
{\epsfig{file=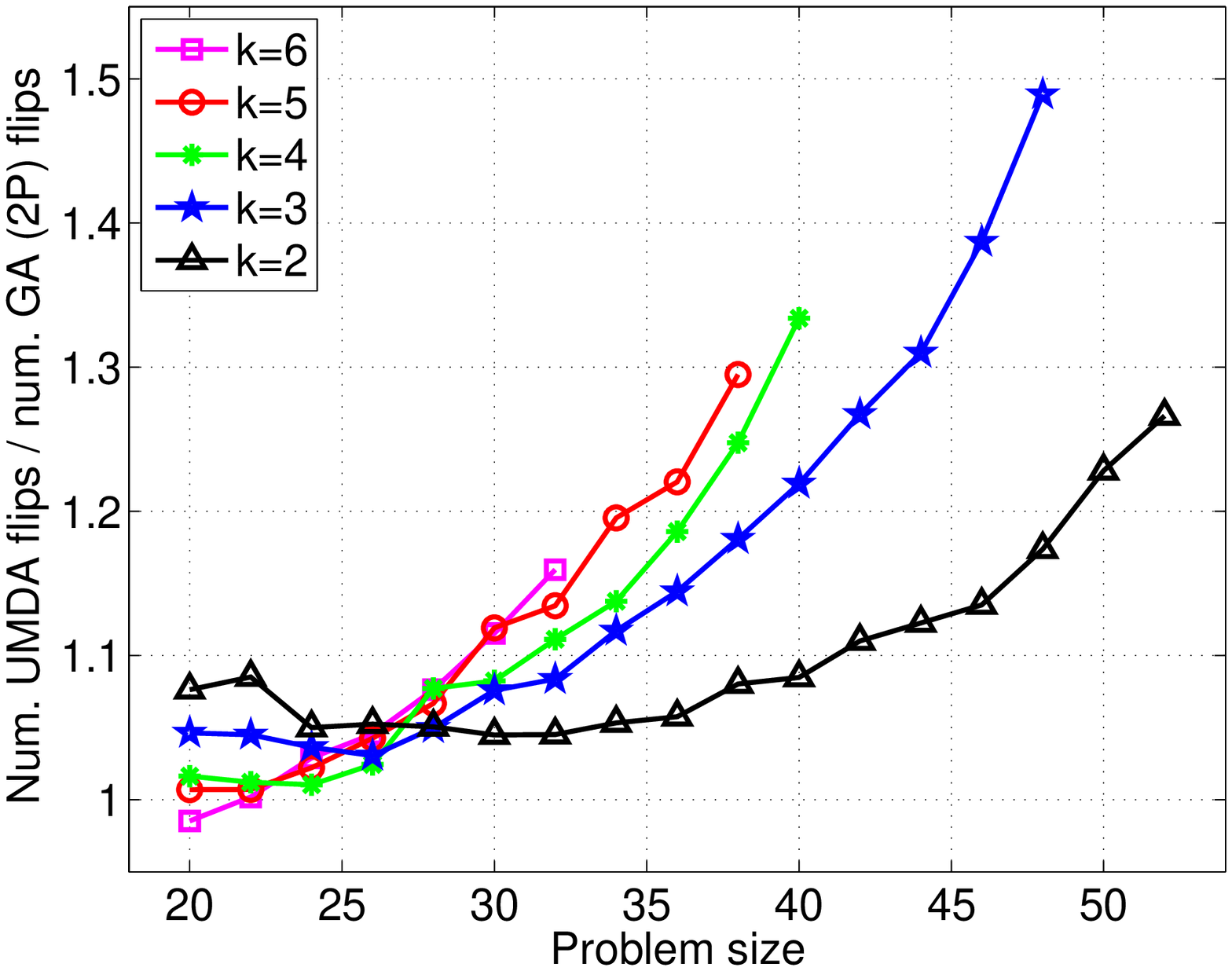,width=0.450\textwidth}}
\caption{Comparison of GA with two-point crossover and UMDA with respect to the number of evaluations and the number of flips. The comparison is visualized by the average ratio of the number of evaluations (number of flips) required by UMDA and the number of evaluations (number of flips) required by GA with two-point crossover. The greater the ratio, the better the performance of GA compared to UMDA. }
\label{fig-compare-ga2p-umda}
\end{figure}

\begin{figure}
\centering
{\epsfig{file=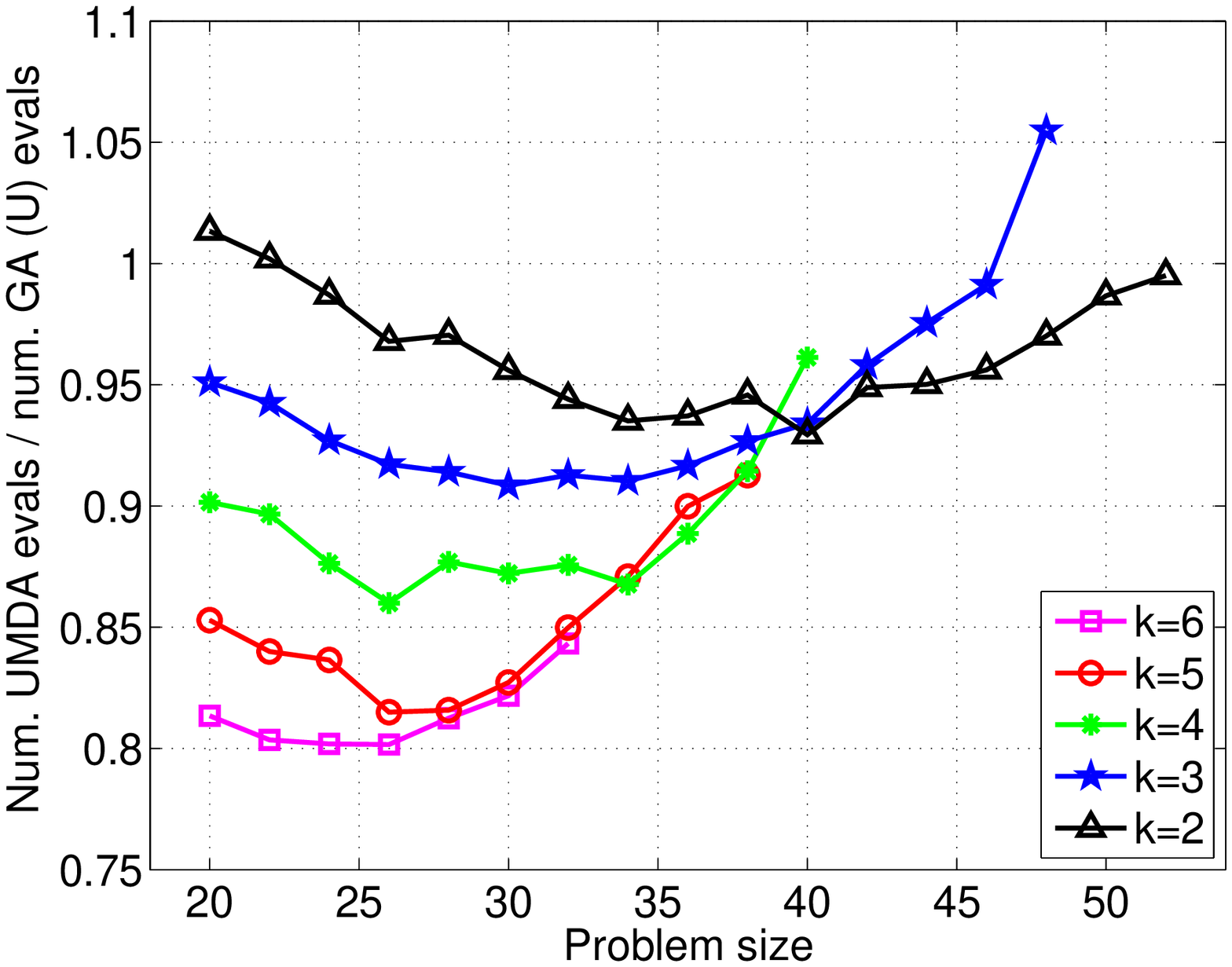,width=0.450\textwidth}}
\hspace*{3ex}
{\epsfig{file=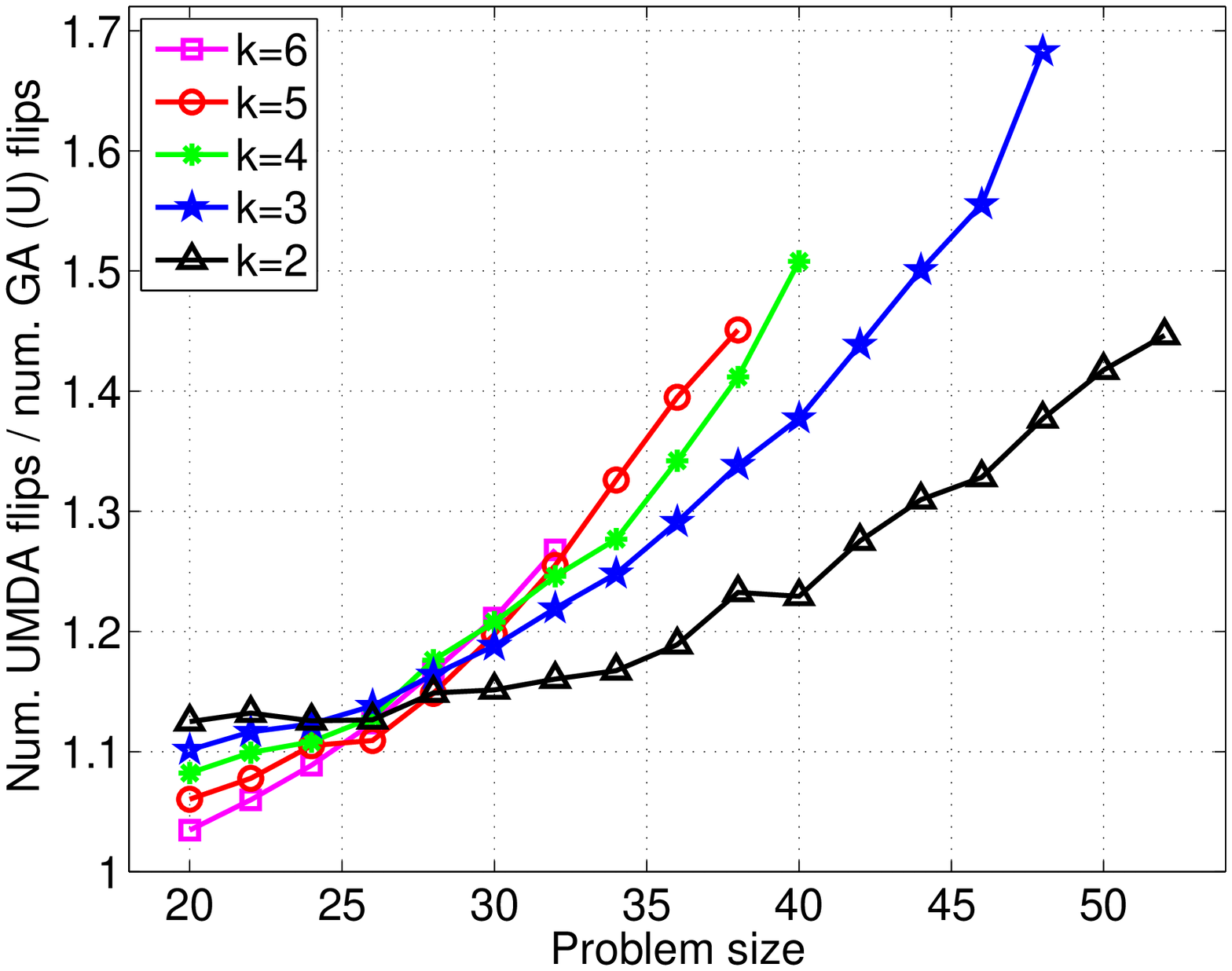,width=0.450\textwidth}}
\caption{Comparison of GA with uniform crossover and UMDA with respect to the number of evaluations and the number of flips. The comparison is visualized by the average ratio of the number of evaluations (number of flips) required by UMDA and the number of evaluations (number of flips) required by GA with uniform crossover. The greater the ratio, the better the performance of GA compared to UMDA.}
\label{fig-compare-gau-umda}
\end{figure}

%==============================================================

\section{Future Work}
\label{section-future-work}
There are several interesting ways of extending the work presented in this paper. First of all, the problem instances generated in this work can be used for analyzing performance of other optimization algorithms and comparing different optimization algorithms on a broad class of problems with tunable difficulty. Second, the class of problems considered in this study can be extended substantially using genetic and evolutionary algorithms with adequate settings for solving instances unsolvable with branch and bound. Although the global optimum would no longer be guaranteed, methods can be devised that still guarantee that the global optimum is found reliably. Finally, other probability distributions for generating NK problem instances can be considered to provide further insights into the difficulty of various classes of NK landscapes and the benefits and costs of using alternative optimization strategies in each of these classes.

%==============================================================

\section{Summary and Conclusions}
\label{section-conclusions}
This paper presented an in-depth empirical performance study of several genetic and evolutionary algorithms on NK landscapes with various values of $n$ and $k$. Specifically, the algorithms considered in this work included the hierarchical Bayesian optimization algorithm (hBOA), the univariate marginal distribution algorithm (UMDA), and the simple genetic algorithm (GA) with bit-flip mutation, and two-point or uniform crossover. Additionally, GA with bit-flip mutation but no crossover was considered. For each value of $n$ and $k$, a large number of NK instances were generated and solved with the branch-and-bound algorithm, which is a complete algorithm that is guaranteed to find the global optimum. Performance of all algorithms was analyzed and compared, and the results were discussed.

The main contributions of this work are summarized in what follows. First of all, NK landscapes represent an important class of test problems and despite that there has been practically no work on using advanced estimation of distribution algorithms (EDAs) on NK landscapes. This work provides many experimental results on one advanced and one simple EDA, and it shows that advanced EDAs can significantly outperform other genetic and evolutionary algorithms on NK landscapes for larger values of $k$. Second, most studies concerned with NK landscapes do not verify the global optimum of the considered problem instances and it is thus often difficult to interpret the results and evaluate their importance. In this study, the global optimum of each instance is verified with the complete branch-and-bound algorithm. Third, while the difficulty of NK landscapes can be expected to vary substantially from instance to instance, most studies presented in the past used only a limited sample of problem instances; here we provide an in-depth study where about 600,000 unique problem instances are considered. Finally, the results in this paper are not based on only one evolutionary algorithm; instead, we consider several qualitatively different evolutionary algorithms, providing insight into the comparison of genetic algorithms and EDAs, as well as into the comparison of the mutation-based and recombination-based evolutionary algorithms.

%==============================================================

\section*{Acknowledgments}
This project was sponsored by the National Science Foundation under CAREER grant ECS-0547013, by the Air Force Office of Scientific Research, Air Force Materiel Command, USAF, under grant FA9550-06-1-0096, and by the University of Missouri in St. Louis through the High Performance Computing Collaboratory sponsored by Information Technology Services, and the Research Award and Research Board programs. 

The U.S.  Government is authorized to reproduce and distribute reprints for government purposes notwithstanding any copyright notation thereon. Any opinions, findings, and conclusions or recommendations expressed in this material are those of the authors and do not necessarily reflect the views of the National Science Foundation, the Air Force Office of Scientific Research, or the U.S. Government. Some experiments were done using the hBOA software developed by Martin Pelikan and David E. Goldberg at the University of Illinois at Urbana-Champaign and most experiments were performed on the Beowulf cluster maintained by ITS at the University of Missouri in St. Louis.

\bibliographystyle{abbrv}

\end{document}